\documentclass{article} 
\usepackage[preprint]{colm2026_conference}

\usepackage{microtype}
\usepackage{hyperref}
\usepackage{url}
\usepackage{booktabs}

\usepackage{graphicx} 
\usepackage{booktabs}       
\usepackage{float}          

\usepackage{xcolor}  
\newcommand{\up}[1]{\textcolor{green!60!black}{$\uparrow$#1}}
\newcommand{\dn}[1]{\textcolor{red!70!black}{$\downarrow$#1}}

\usepackage{lineno}

\usepackage{graphicx}
\usepackage{subcaption}
\usepackage{pifont}  
\newcommand{\cmark}{\ding{51}}
\newcommand{\xmark}{\ding{55}}

\definecolor{darkblue}{rgb}{0, 0, 0.5}
\hypersetup{colorlinks=true, citecolor=darkblue, linkcolor=darkblue, urlcolor=darkblue}

\usepackage{colortbl}

\definecolor{azure(web)(azuremist)}{rgb}{0.94, 1.0, 1.0}
\definecolor{oldlace}{rgb}{0.99, 0.96, 0.9}
\definecolor{pearl}{rgb}{0.94, 0.92, 0.84}
\definecolor{seashell}{rgb}{1.0, 0.96, 0.93}
\definecolor{silver}{rgb}{0.75, 0.75, 0.75}
\definecolor{platinum}{rgb}{0.9, 0.89, 0.89}
\definecolor{almond}{rgb}{0.94, 0.87, 0.8}
\definecolor{lightskyblue}{RGB}{173, 216, 230}
\definecolor{mintcream}{rgb}{0.96, 1.0, 0.98}
\definecolor{honeydew}{rgb}{0.94, 1.0, 0.94}
\definecolor{lavenderblush}{rgb}{1.0, 0.94, 0.96}
\definecolor{aliceblue}{rgb}{0.94, 0.97, 1.0}
\definecolor{mistyrose}{rgb}{1.0, 0.89, 0.88}
\definecolor{linen}{rgb}{0.98, 0.94, 0.9}
\definecolor{lemonchiffon}{rgb}{1.0, 0.98, 0.8}

\usepackage{soul}
\usepackage{xspace}
\usepackage[most]{tcolorbox}

\definecolor{lightred}{RGB}{255, 204, 204}

\newtcolorbox{llmbox}[2][]{
  breakable,
  enhanced,
  colback=white,
  colframe=black!70,
  colbacktitle=black!70,
  coltitle=white,
  fonttitle=\bfseries\small,
  title={#2},
  left=8pt, right=8pt, top=4pt, bottom=4pt,
  arc=2pt,
  #1
}
\newcommand{\tvar}[1]{\textcolor{teal}{\texttt{[#1]}}}
\usepackage{enumitem}

\newcommand{\OURSBENCH}{AgentSearchBench}

\def \TAB {Table}
\def \FIG {Figure}

\title{AgentSearchBench: A Benchmark for AI Agent Search \\ in the Wild}

\author{Bin Wu\thanks{Equal Contribution}
\,\thanks{Corresponding author: \href{mailto:bin.wu.23@ucl.ac.uk}{bin.wu.23@ucl.ac.uk}}, 
\ \ Arastun Mammadli$^{*}$, 
\ Xiaoyu Zhang, 
\ Emine Yilmaz \\
Centre for Artificial Intelligence, University College London
}

\usepackage{amsmath,amsfonts,bm}

\def\eqref#1{equation~\ref{#1}}

\def\1{\bm{1}}

\DeclareMathAlphabet{\mathsfit}{\encodingdefault}{\sfdefault}{m}{sl}
\SetMathAlphabet{\mathsfit}{bold}{\encodingdefault}{\sfdefault}{bx}{n}

\begin{document}

\ifcolmsubmission
\linenumbers
\fi

\maketitle

\begin{abstract}
The rapid growth of AI agent ecosystems is transforming how complex tasks are delegated and executed, creating a new challenge of identifying suitable agents for a given task. Unlike traditional tools, agent capabilities are often compositional and execution-dependent, making them difficult to assess from textual descriptions alone. However, existing research and benchmarks typically assume well-specified functionalities, controlled candidate pools, or only executable task queries, leaving realistic agent search scenarios insufficiently studied. We introduce AgentSearchBench, a large-scale benchmark for agent search in the wild, built from nearly 10,000 real-world agents across multiple providers. The benchmark formalizes agent search as retrieval and reranking problems under both executable task queries and high-level task descriptions, and evaluates relevance using execution-grounded performance signals. Experiments reveal a consistent gap between semantic similarity and actual agent performance, exposing the limitations of description-based retrieval and reranking methods. We further show that lightweight behavioral signals, including execution-aware probing, can substantially improve ranking quality, highlighting the importance of incorporating execution signals into agent discovery. Our code is available at \href{https://github.com/Bingo-W/AgentSearchBench}{https://github.com/Bingo-W/AgentSearchBench}.

\end{abstract}

\section{Introduction}
The rapid emergence of AI agentic systems is reshaping how humans accomplish complex tasks by enabling execution to be delegated to autonomous agents across a wide range of domains \citep{DBLP:journals/corr/abs-2508-07407, DBLP:journals/tmlr/GaoGHHJLLQQRWWXZZZXFZLQ26}. Modern agents can reason, plan, and interact with external tools and services to complete multi-step objectives \citep{DBLP:journals/corr/abs-2402-02716, DBLP:journals/corr/abs-2504-19678, DBLP:journals/csur/QinHLCDCZZHXHFSWQTZLSXZ25}. This progress has led to a rapidly expanding ecosystem of agentic components, ranging from general-purpose assistants to highly specialized task-oriented modules. As humans increasingly rely on agents developed by diverse third-party providers, a fundamental challenge arises: \textit{how can suitable agents be reliably identified and selected for a given task?} Addressing this challenge is critical not only for end users seeking effective task completion, but also for developers and orchestration systems aiming to compose scalable and robust agentic workflows \citep{DBLP:journals/corr/abs-2411-04468, hu2025owl}.

However, identifying suitable agents is inherently challenging. Compared to traditional tools whose functionality is typically scoped to specific operations \citep{shi2025retrieval}, agent capabilities are often more compositional and execution-dependent, making them difficult to assess without observing task outcomes. As a result, textual descriptions provide only a partial signal of real competence \citep{DBLP:conf/iclr/QuDWCWY0W25, wu2025joint, fang2025play2prompt}: agents with similar descriptions may perform differently in practice, while semantically dissimilar agents can achieve comparable results. This semantic–performance misalignment is further amplified in large and open agent ecosystems, where overlapping functionalities and non-uniform description formats make capability comparison difficult \citep{yuan2025easytool}. Consequently, agent search is fundamentally more complex than conventional tool retrieval or model selection.

Despite growing interest in agentic systems, existing research and benchmarks have not yet provided a realistic setting for studying agent search (in \TAB~\ref{tab:benchmark_comparison}). Prior work on tool retrieval and related benchmarks primarily assumes that functionality can be inferred from structured descriptions or well-specified interfaces \citep{DBLP:conf/iclr/QinLYZYLLCTQZHT24, shi2025retrieval}, which does not capture the compositional and execution-dependent nature of agent capabilities. Meanwhile, recent studies on automated agentic system design typically evaluate methods in small-scale or controlled environments where candidate agents are clearly differentiated \citep{DBLP:conf/iclr/ShangLZMLXL25, yuan2025automated}. Such assumptions differ substantially from open ecosystems, where many agents exhibit overlapping capabilities and must be selected under uncertainty. Furthermore, existing tool and agent selection benchmarks largely focus on executable task queries with predefined inputs and outputs, whereas real-world agent discovery often begins from high-level task descriptions that are not directly executable. As a result, performance-grounded agent search in realistic open ecosystems remains insufficiently studied.

To bridge these gaps, we introduce \OURSBENCH{}, a large-scale benchmark for agent search built from nearly 10,000 real-world agents. \OURSBENCH{} captures the diversity of open agent ecosystems by including agents from different providers with varying description styles, capability granularity, and functional overlap. Built on this resource, we formalize agent search as retrieval and reranking problems under both executable task queries and high-level task descriptions. Crucially, agent relevance is defined using execution-grounded performance signals rather than textual similarity. We further develop a scalable evaluation pipeline that generates task instances and converts execution outcomes into fine-grained relevance annotations for both retrieval and ranking assessment.

Through extensive benchmarking experiments, we reveal a consistent gap between semantic similarity and actual task performance, providing empirical evidence for the execution-dependent nature of agent capabilities. Retrieval and reranking methods that rely primarily on matching task descriptions with agent documentation often fail to surface high-performing agents, particularly when search begins from high-level task descriptions where capability requirements are implicit. To better understand this limitation, we further study execution-aware probing, which augments description-based ranking with lightweight behavioral signals obtained from agent execution. Results show that even limited probing can substantially improve ranking quality, highlighting the importance of incorporating execution signals into realistic agent discovery pipelines.

Our contributions can be summarized as follows:
(1) We formulate agent search as a new retrieval and ranking problem under execution-dependent capability uncertainty.
(2) We construct \OURSBENCH{}, a large-scale benchmark with nearly 10,000 real-world agents, supporting both executable task queries and high-level task descriptions under an execution-grounded evaluation framework.
(3) We provide extensive empirical analysis revealing a substantial semantic–performance gap and demonstrate the effectiveness of lightweight behavioral probing for improving agent ranking.

\begin{table}[!t]
    \centering
    \resizebox{0.9\textwidth}{!}{\begin{tabular}{c|ccccc}
        \toprule
        & Agent/LLM/Tool & \#Candidates &Realistic  & Task Type\\
        \midrule
        ToolBench \citep{DBLP:conf/iclr/QinLYZYLLCTQZHT24} &Tool&16,464&\cmark& Exec.  \\
         ToolRet \citep{shi2025retrieval}&Tool&43,215&\cmark& Exec.  \\
        TREC 2025 \citep{trecmllm2025} &LLM&1,131&\cmark& Exec.  \\
        AgentSquare \citep{DBLP:conf/iclr/ShangLZMLXL25} & Agent & 16 &\xmark& Exec.  \\
        OKC Bench \citep{yuan2025automated} & Agent & 127 &\xmark& Exec.  \\
        \hline
         \OURSBENCH{}&Agent&9,759&\cmark& Exec. / Non-exec. \\
         \bottomrule
    \end{tabular}}
    \caption{Comparison of \OURSBENCH{} with tool and agent retrieval benchmarks. ``Realistic'' indicates whether agents/tools are sourced from real-world platforms. ``Task Type'' indicates support for executable (Exec.) or non-executable (Non-exec.) task specifications.}
    \vspace{-1em}
    \label{tab:benchmark_comparison}
\end{table}

\section{Related Work}
\textbf{Agentic Systems and Orchestration. } Recent advances in agentic systems have enabled autonomous agents to solve complex tasks through reasoning, planning, and tool interaction \citep{DBLP:journals/corr/abs-2402-02716, DBLP:journals/corr/abs-2504-19678, DBLP:journals/csur/QinHLCDCZZHXHFSWQTZLSXZ25}. As agent ecosystems rapidly expand, a key challenge is how to select suitable agents for a given task. Existing work focuses on agent design and orchestration, including multi-agent collaboration \citep{DBLP:journals/corr/abs-2411-04468, hu2025owl} and workflow composition frameworks \citep{yue-etal-2025-masrouter, DBLP:conf/iclr/ShangLZMLXL25, yuan2025automated}, typically assuming a predefined and limited set of candidates. However, in open ecosystems with many overlapping agents \citep{yuan2025easytool, wu2025joint, zhang2025routerr}, agent selection must be performed under significant capability uncertainty, motivating agent search as a distinct problem.

\textbf{Tool Retrieval and Selection. } Existing works on tool retrieval and selection \citep{DBLP:conf/iclr/QinLYZYLLCTQZHT24, shi2025retrieval} aim to identify suitable tools for task execution. These methods typically retrieve tools based on textual descriptions or structured schemas \citep{DBLP:journals/corr/abs-2602-05366, lu2025tools, qu2024towards}, and are primarily evaluated on executable task queries with predefined inputs and outputs \citep{DBLP:conf/iclr/QinLYZYLLCTQZHT24, qu2024towards}. While effective in such settings, these assumptions do not capture real-world agent search, which often begins from high-level and non-executable task descriptions. Moreover, agent capabilities are compositional, inconsistently documented, and execution-dependent, making textual similarity insufficient for assessing suitability. Recent work explores improving tool representations via schema unification or execution signals \citep{yuan2025easytool, DBLP:conf/iclr/QuDWCWY0W25, wu2025joint}, but mainly focuses on tool usage rather than discovery over large candidate pools. In contrast, we formulate agent search as a performance-grounded problem that supports both executable queries and high-level task descriptions.

\textbf{Information Retrieval and Learning-to-Rank. } Information retrieval and learning-to-rank provide the foundation for modeling agent search \citep{robertson2009probabilistic, behnamghader2024llm2vec}. Existing methods estimate relevance using textual similarity or annotated labels \citep{craswell2021ms, craswell2025overview}. However, they assume relevance is static and observable without interaction, which does not hold for agents. In agent search, relevance is inherently execution-dependent, requiring evaluation through task performance. We therefore extend retrieval and ranking to incorporate execution-grounded relevance signals.

\section{Problem Formulation}

\subsection{Agent Search Problem}

Agent search aims to retrieve and rank suitable agents from a large candidate repository given a user task. Let $\mathcal{T}$ denote the task specification, and let $\mathcal{C} = \{a_1, a_2, \dots, a_n\}$ denote the candidate agent repository. Each agent is represented by descriptive documentation and, when available, an executable interface. An agent search system defines a scoring function $f(a,\mathcal{T})$ that estimates the relevance of an agent $a \in \mathcal{C}$ to task $\mathcal{T}$, and produces a ranked list
\begin{align}
\mathcal{O}_{\mathrm{ranked}} = \operatorname{argsort}_{a \in \mathcal{C}} f(a,\mathcal{T}).
\end{align}
Agent search involves two objectives: (1) retrieving top-k agents capable of solving the task, and (2) ranking them according to task performance quality. Unlike traditional information retrieval, where relevance is determined by static content matching, agent search requires assessing functional capability through task execution.

\subsection{Task Query and Task Description}

Agent search operates under different levels of task specification. We consider two types of inputs: executable task queries and high-level task descriptions.

\textbf{Task Query. }
A task query $\mathcal{T}_q$ is a concrete and executable instruction that can be directly evaluated by running an agent. Task queries include both single-agent tasks and multi-agent tasks composed of multiple executable subtasks \citep{DBLP:conf/iclr/QinLYZYLLCTQZHT24, shi2025retrieval}. Given a task query, the goal is to identify agents that can successfully complete the task and rank them based on their execution performance.

\textbf{Task Description. }
In many realistic scenarios, users provide high-level goals that are not directly executable. We denote such inputs as task descriptions $\mathcal{T}_d$. To evaluate agent capability under these settings, each task description is associated with a set of executable task queries, $\mathcal{Q}(\mathcal{T}_d) = \{\mathcal{T}_{q_1}, \dots, \mathcal{T}_{q_m}\}$,
which instantiate the high-level goal under different concrete scenarios. Agent relevance is then determined based on performance across $\mathcal{Q}(\mathcal{T}_d)$, enabling evaluation of consistent capability rather than success on a single task instance.

\subsection{Task-Performance-Based Relevance}

Relevance in agent search is defined based on task execution performance. For an executable task query $\mathcal{T}_q$, we define the relevance of an agent $a$ using a task completion score $y(a,\mathcal{T}_q) = \mathrm{E}(a,\mathcal{T}_q)$, where $\mathrm{E}$ evaluates the quality of the agent’s response (e.g., via LLM-as-a-judge). For a task description $\mathcal{T}_d$, relevance cannot be evaluated directly. Instead, we aggregate performance over its associated queries:
\begin{align}
    y(a,\mathcal{T}_d) = \frac{1}{|\mathcal{Q}(\mathcal{T}_d)|} \sum_{\mathcal{T}_q \in \mathcal{Q}(\mathcal{T}_d)} y(a,\mathcal{T}_q).
\end{align}
This formulation measures agent capability as consistent performance across multiple task instances, rather than relying on textual similarity or single-task outcomes.

In practice, observed execution performance may not always align with documented capabilities. When an agent successfully solves a task that is not supported by its documentation, we treat the success as potentially less reliable than that of an agent whose documented functionality is consistent with its observed performance. Accordingly, we incorporate \textbf{documentation–performance alignment} as an auxiliary signal when constructing ranking labels, so that relevance reflects both task success and the reliability of that success.

\section{AgentSearchBench: A Benchmark for Agent Search}

\begin{figure*}[t]
\centering
\includegraphics[width=\linewidth]{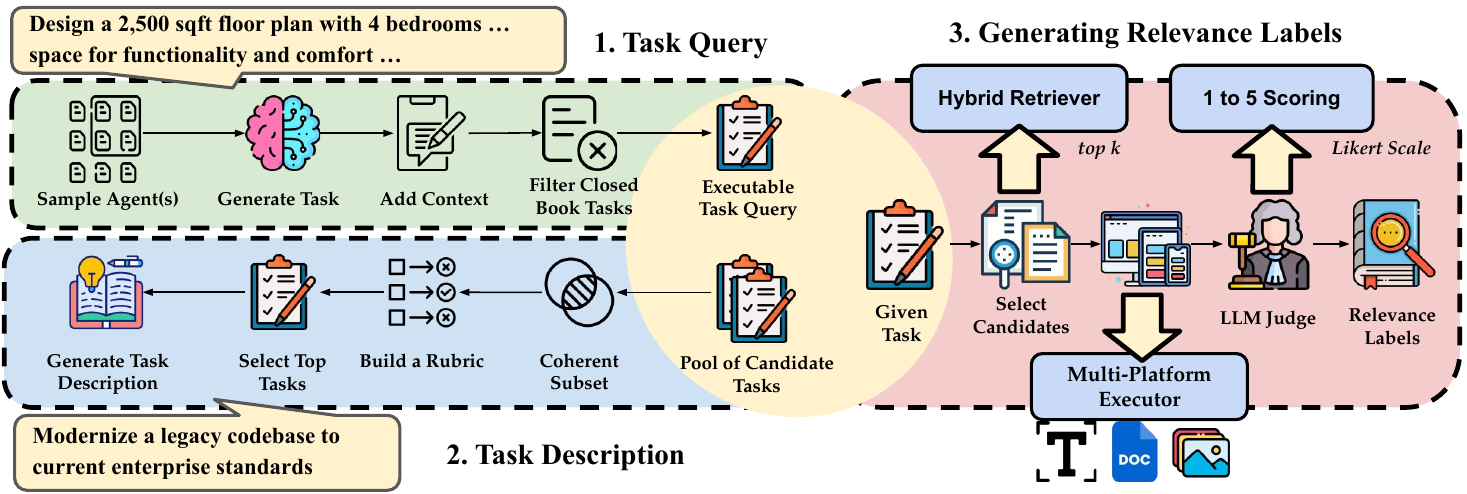}
\caption{Task and Relevance Label Generation Pipeline of \OURSBENCH{}.}
\label{fig:pipeline}
\end{figure*}

We construct \OURSBENCH{} from a large-scale collection of real-world agents and generate tasks through a hierarchical process. We first create executable task queries, then derive high-level task descriptions grounded in query-level evidence. Relevance is obtained via execution-based evaluation and converted into retrieval and ranking labels. The overall pipeline is shown in \FIG~\ref{fig:pipeline}.

\subsection{Large-Scale Realistic Agent Collection}
We build a repository of nearly 10k agents collected from public platforms, including GPT store\footnote{https://chatgpt.com/gpts}, Google Cloud Marketplace\footnote{https://cloud.google.com/marketplace}, and AgentAI Platform\footnote{https://agent.ai/}. By sourcing agents from real-world ecosystems rather than synthesizing them, AgentBase captures practical challenges such as capability overlap and inconsistent documentation, enabling realistic evaluation of agent search.

\subsection{Task Query Construction}

We synthesize executable task queries from agent documentation following document-grounded task generation \citep{DBLP:conf/iclr/QinLYZYLLCTQZHT24}. To reduce evaluation cost, we retrieve a candidate set of top-$K$ agents using a hybrid scoring function that combines BM25 (lexical) \citep{robertson2009probabilistic}, BGE (semantic) \citep{xiao2024cpack}, and ToolRet (tool-aware) retrieval \citep{shi2025retrieval}:
\begin{align}
s(a,\mathcal{T}_q)=\alpha\,s_{\text{lexical}}+\beta\,s_{\text{semantic}}+\gamma\,s_{\text{tool}}.
\end{align}
Retrieved agents are executed on each query and evaluated using a 5-point LLM-as-judge \citep{gu2024survey}. We filter out degenerate queries where no agent succeeds, or all agents succeed.

Multi-agent queries are constructed by composing executable subtasks from capability-aligned clusters. We retain a composed query only if it semantically entails all subtasks via natural language inference (NLI) \citep{wu2025natural}:
\begin{align}
\mathrm{Entail}(\mathcal{T}_q^{\text{multi}},\mathcal{T}_q^{(i)})=1,\ \forall i.
\end{align}

\subsection{Task Description Construction}

We construct task descriptions by abstracting high-level objectives from clusters of semantically related queries. Given query clusters $\{\mathcal{C}_1,\dots,\mathcal{C}_M\}$, we remove outliers and generate a description $\mathcal{T}_d$ for each cluster using LLMs.

To associate executable queries with each description, we apply a rubric-based judge \citep{sharma2026researchrubrics} that evaluates the relevance of each candidate query $\mathcal{T}_q$ to $\mathcal{T}_d$ from multiple aspects. Formally, let $\mathbf{r}(\mathcal{T}_d,\mathcal{T}_q)=\big(r_1(\mathcal{T}_d,\mathcal{T}_q),\dots,r_D(\mathcal{T}_d,\mathcal{T}_q)\big)$ denote the aspect-wise relevance scores. For each aspect $d$, we select the top-2 queries according to $r_d$, and construct the associated query set as $\mathcal{Q}(\mathcal{T}_d)=\bigcup_{d=1}^{D}\text{Top2}_{\mathcal{T}_q}\, r_d(\mathcal{T}_d,\mathcal{T}_q)$,
resulting in 10 queries when $D=5$. To ensure reliable evaluation, we re-evaluate high-performing agents on missing subtasks and filter inconsistent task descriptions.

\subsection{Relevance Annotation}

Relevance is derived from execution-based performance using a 5-point LLM-as-judge. For retrieval, we convert scores into binary labels:
\begin{align}
\mathrm{rel}(a,\mathcal{T}_q)=\mathbf{1}(y(a,\mathcal{T}_q)\geq 4).
\end{align}

For multi-agent queries and task descriptions, we define graded relevance based on subtask completion:
\begin{align}
r(a,\mathcal{T})=\frac{1}{|\mathcal{S}(\mathcal{T})|}\sum_{\mathcal{T}_q\in\mathcal{S}(\mathcal{T})}\mathrm{rel}(a,\mathcal{T}_q).
\end{align}

To account for documentation–performance misalignment, agents that successfully complete tasks without corresponding documented capability are assigned discounted relevance scores (e.g., 0.5). These signals are used to construct golden rankings for evaluation.

\section{Benchmark Statistic}

\begin{figure*}[t]
\centering

\begin{subfigure}[b]{0.34\textwidth}
\vspace{0pt}
\centering
\small
\resizebox{\linewidth}{!}{%
\begin{tabular}{l r}
\toprule
\multicolumn{2}{l}{\textbf{Statistics}} \\
\midrule
\# Total Agents & 9,759 \\
\# Total Task & 3,211 \\
\quad - \# Single-Agent Task Query & 2,452  \\
\quad - \# Multi-Agent Task Query & 500  \\
\quad - \# Task Description & 259 \\
Avg. query per description & 10 \\
Avg. evaluated agents per query & 20 \\
\# Total executions & 66,740  \\
\bottomrule
\end{tabular}%
}
\caption{Overview statistics.}
\end{subfigure}
\hfill
\begin{subfigure}[b]{0.34\textwidth}
\vspace{0pt}
\centering
\includegraphics[width=\linewidth]{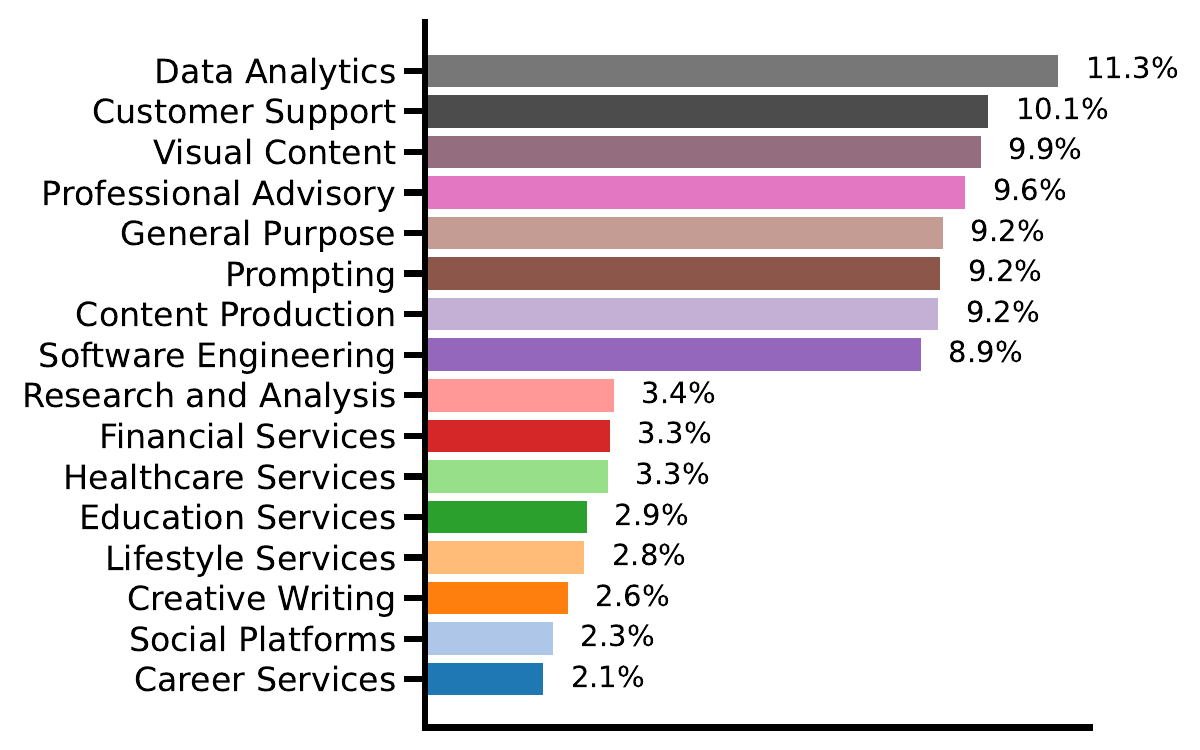}
\caption{Agent diversity.}
\end{subfigure}
\hfill
\begin{subfigure}[b]{0.26\textwidth}
\vspace{0pt}
\centering
\includegraphics[width=\linewidth]{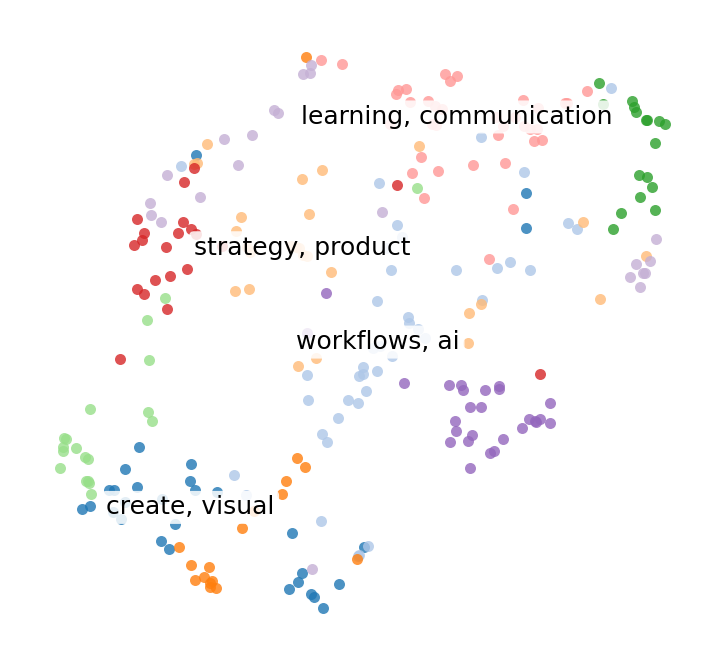}
\caption{Task diversity.}
\end{subfigure}
\vspace{-0.5em}
\caption{Benchmark statistics and semantic diversity of \OURSBENCH{}.}
\vspace{-0.5em}
\label{fig:benchmark_stats}
\end{figure*}

\begin{figure*}[t]
\begin{subfigure}[b]{0.3\textwidth}
\centering
\includegraphics[width=\linewidth]{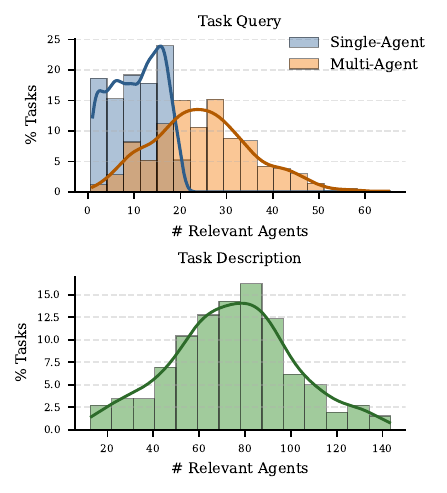}
\caption{Relevant agents per query.}
\end{subfigure}
\hfill
\begin{subfigure}[b]{0.3\textwidth}
\centering
\includegraphics[width=\linewidth]{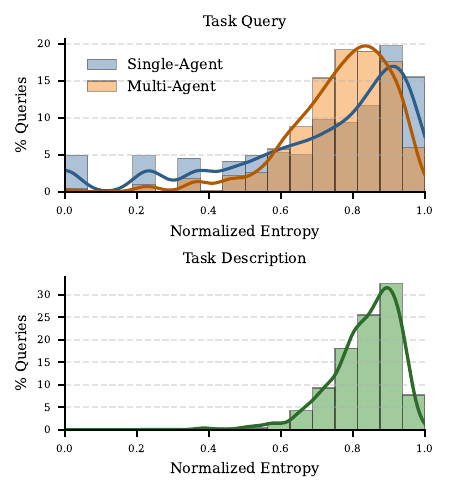}
\caption{Score entropy.}
\end{subfigure}
\hfill
\begin{subfigure}[b]{0.3\textwidth}
\centering
\includegraphics[width=\linewidth]{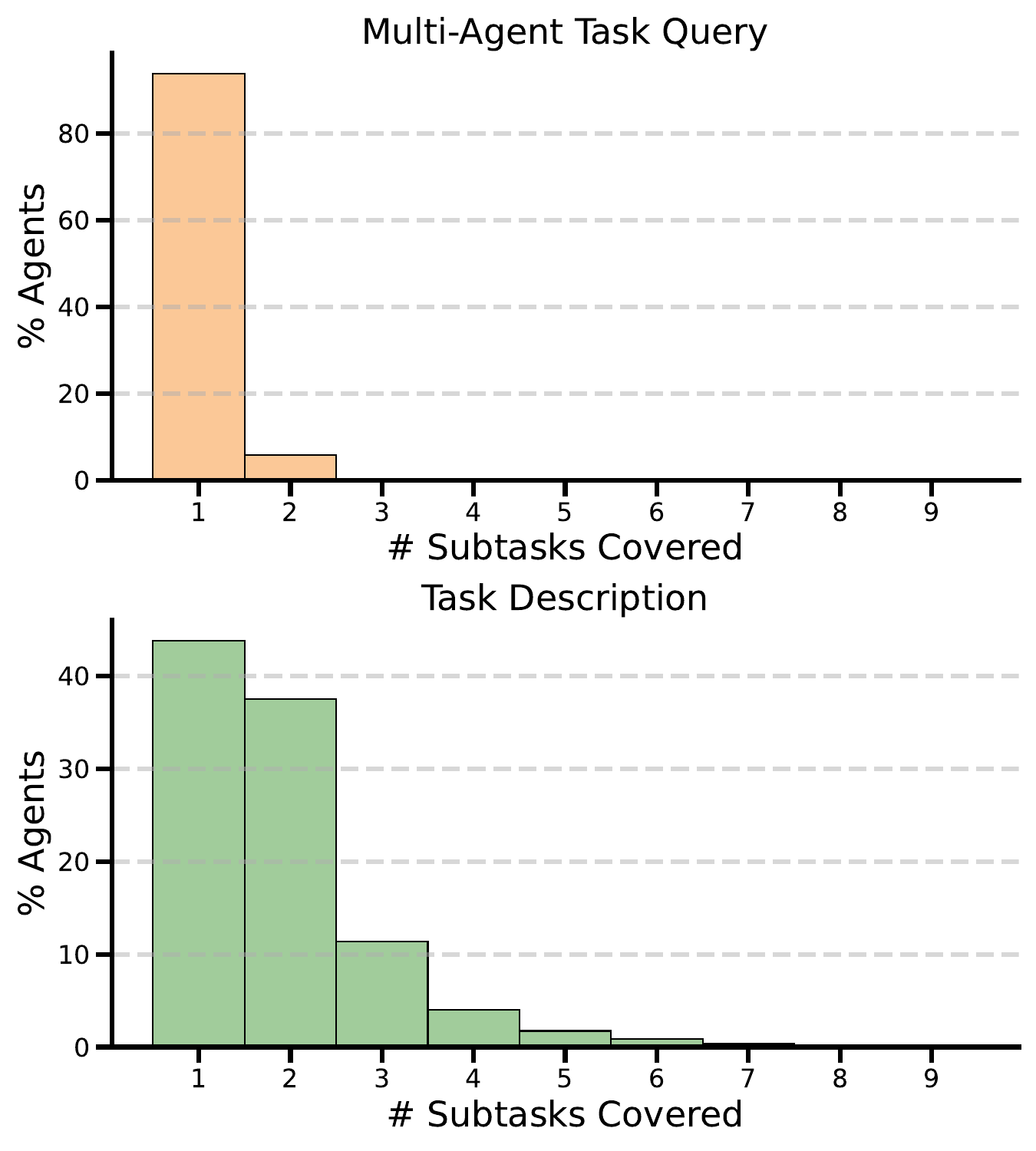}
\caption{Subtask coverage.}
\end{subfigure}
\vspace{-0.5em}

\caption{Difficulty of Task Query and Task Description of \OURSBENCH{}.}
\vspace{-0.5em}
\label{fig:benchmark_stats_difficulty}
\end{figure*}

We summarize the scale of \OURSBENCH{} in \FIG~\ref{fig:benchmark_stats}. The benchmark contains nearly \textbf{9,760} agents collected from multiple open platforms, among which \textbf{7,867} provide executable interfaces. We construct \textbf{2,952} executable task queries and \textbf{259} task descriptions, each associated with an average of \textbf{10} queries. Each query is evaluated on the top-\textbf{20} retrieved agents, resulting in \textbf{66,740} execution runs. These statistics highlight the scale and execution-centric design of \OURSBENCH{}.

\textbf{Diversity and Difficulty. }
As shown in \FIG~\ref{fig:benchmark_stats}(b–c), both agents and tasks exhibit broad and long-tailed semantic distributions, reflecting diverse capability coverage. \FIG~\ref{fig:benchmark_stats_difficulty}(a) shows that many queries have multiple relevant agents, making retrieval alone insufficient. Meanwhile, the entropy distribution in \FIG~\ref{fig:benchmark_stats_difficulty}(b) indicates substantial performance variance across agents, motivating fine-grained reranking. Finally, \FIG~\ref{fig:benchmark_stats_difficulty}(c) shows that agents typically cover only a subset of subtasks, highlighting partial and overlapping capabilities. Overall, \OURSBENCH{} provides a realistic and challenging setting for performance-grounded agent search.

\section{Agent Search Evaluation}

\begin{table}[!t]
\centering
\renewcommand{\arraystretch}{1.25}
\resizebox{\textwidth}{!}{%
\begin{tabular}{lccccccccccccc}
\toprule
\textbf{Model} & \textbf{Type} & \multicolumn{3}{c}{\textbf{NDCG}} & \multicolumn{3}{c}{\textbf{Precision}} & \multicolumn{3}{c}{\textbf{Recall}} & \multicolumn{3}{c}{\textbf{Completeness}}\\
\cline{3-14}
 & & \textbf{@5} & \textbf{@10} & \textbf{@20} & \textbf{@5} & \textbf{@10} & \textbf{@20} & \textbf{@5} & \textbf{@10} & \textbf{@20} & \textbf{@5} & \textbf{@10} & \textbf{@20} \\
\midrule
\rowcolor{lavenderblush}\multicolumn{14}{c}{\textit{Task Query}} \\
\hline
SPLADE v2 & Sparse & 4.09 & 3.49 & 3.48 & 4.09 & 3.19 & 2.29 & 0.35 & 1.35 & 3.72 & 2.90 & 9.21 & 18.93 \\
BM25 & Sparse & \cellcolor{aliceblue}\textbf{32.41} & \cellcolor{aliceblue}\textbf{26.73} & \cellcolor{aliceblue}\textbf{22.68} & \cellcolor{aliceblue}\textbf{32.41} & \cellcolor{aliceblue}\textbf{24.56} & \cellcolor{aliceblue}\textbf{14.07} & \cellcolor{aliceblue}\textbf{2.62} & \cellcolor{aliceblue}\textbf{9.33} & \cellcolor{aliceblue}\textbf{21.36} & \cellcolor{aliceblue}\textbf{20.81} & \cellcolor{aliceblue}\textbf{38.10} & \cellcolor{aliceblue}\textbf{51.38} \\
\hline
ColBERT v2 & Dense & 27.71 & 23.07 & 20.47 & 27.71 & 21.14 & 13.44 & 2.12 & 7.94 & 19.68 & 16.52 & 33.89 & 43.95 \\
Contriever MS-Marco & Dense & 20.75 & 17.43 & 16.72 & 20.75 & 16.11 & 11.12 & 1.72 & 6.34 & 17.07 & 15.04 & 30.73 & 45.11 \\
GTR-T5 Base & Dense & 23.20 & 19.50 & 17.55 & 23.20 & 18.05 & 11.51 & 1.89 & 6.87 & 16.92 & 16.00 & 32.09 & 44.63 \\
MiniLM-L6 v2 & Dense & 29.02 & 25.28 & 22.46 & 29.02 & 23.57 & 14.97 & 2.29 & 8.63 & 21.33 & 18.23 & 36.52 & 51.52 \\
BGE-Large v1.5 & Dense & \cellcolor{aliceblue}\textbf{31.78} & \cellcolor{aliceblue}\textbf{28.24} & \cellcolor{aliceblue}\textbf{26.14} & \cellcolor{aliceblue}\textbf{31.78} & \cellcolor{aliceblue}\textbf{26.48} & \cellcolor{aliceblue}\textbf{17.64} & \cellcolor{aliceblue}\textbf{2.48} & \cellcolor{aliceblue}\textbf{10.07} & \cellcolor{aliceblue}\textbf{25.49} & \cellcolor{aliceblue}\textbf{20.18} & \cellcolor{aliceblue}\textbf{38.61} & \cellcolor{aliceblue}\textbf{52.52} \\
\hline
COLT ToolLens & Tool & 9.35 & 8.12 & 7.91 & 9.35 & 7.59 & 5.47 & 0.81 & 2.92 & 7.85 & 6.95 & 18.07 & 30.61 \\
COLT ToolBench & Tool & 16.25 & 13.84 & 12.28 & 16.25 & 12.81 & 8.00 & 1.30 & 4.82 & 12.13 & 11.74 & 25.39 & 38.66 \\
Tool-Embed & Tool & 34.02 & 28.40 & 25.67 & 34.02 & 26.37 & 17.23 & 2.50 & 9.77 & 24.46 & 20.22 & 38.14 & 53.91 \\
ToolRet & Tool & \cellcolor{aliceblue}\textbf{37.52} & \cellcolor{aliceblue}\textbf{31.73} & \cellcolor{aliceblue}\textbf{28.87} & \cellcolor{aliceblue}\textbf{37.52} & \cellcolor{aliceblue}\textbf{29.36} & \cellcolor{aliceblue}\textbf{19.36} & \cellcolor{aliceblue}\textbf{2.85} & \cellcolor{aliceblue}\textbf{10.97} & \cellcolor{aliceblue}\textbf{27.80} & \cellcolor{aliceblue}\textbf{21.91} & \cellcolor{aliceblue}\textbf{42.18} & \cellcolor{aliceblue}\textbf{57.53} \\
\hline
E5-Mistral 7B & Decoder-only & 19.57 & 16.31 & 15.29 & 19.57 & 14.99 & 9.77 & 1.60 & 6.05 & 15.62 & 15.27 & 30.71 & 44.49 \\
Qwen-Embedding 8B & Decoder-only & \cellcolor{aliceblue}\textbf{25.25} & \cellcolor{aliceblue}\textbf{22.59} & \cellcolor{aliceblue}\textbf{21.43} & \cellcolor{aliceblue}\textbf{25.25} & \cellcolor{aliceblue}\textbf{21.32} & \cellcolor{aliceblue}\textbf{14.83} & \cellcolor{aliceblue}\textbf{1.94} & \cellcolor{aliceblue}\textbf{7.81} & \cellcolor{aliceblue}\textbf{21.14} & \cellcolor{aliceblue}\textbf{16.15} & \cellcolor{aliceblue}\textbf{33.17} & \cellcolor{aliceblue}\textbf{48.33} \\
\midrule
\rowcolor{oldlace}\multicolumn{14}{c}{\textit{Task Description}} \\
\hline
SPLADE v2 & Sparse & 12.02 & 8.24 & 6.93 & 12.02 & 7.50 & 6.11 & 0.35 & 0.83 & 2.49 & 0.00 & 0.00 & \cellcolor{aliceblue}\textbf{0.48} \\
BM25 & Sparse & \cellcolor{aliceblue}\textbf{16.35} & \cellcolor{aliceblue}\textbf{13.42} & \cellcolor{aliceblue}\textbf{10.78} & \cellcolor{aliceblue}\textbf{16.35} & \cellcolor{aliceblue}\textbf{12.50} & \cellcolor{aliceblue}\textbf{9.25} & \cellcolor{aliceblue}\textbf{0.46} & \cellcolor{aliceblue}\textbf{1.28} & \cellcolor{aliceblue}\textbf{3.69} & 0.00 & 0.00 & \cellcolor{aliceblue}\textbf{0.48} \\
\hline
ColBERT v2 & Dense & 15.38 & 13.19 & 12.80 & 15.38 & 12.31 & 12.19 & 0.27 & 1.14 & 5.12 & 0.00 & 0.48 & 1.92 \\
Contriever MS-Marco & Dense & 15.38 & 15.03 & 14.22 & 15.38 & 14.81 & 13.37 & 0.31 & 1.49 & 5.70 & 0.00 & 0.48 & 1.92 \\
GTR-T5 Base & Dense & 15.87 & 15.03 & 12.96 & 15.87 & 14.52 & 11.92 & 0.34 & 1.56 & 4.93 & 0.00 & 0.00 & 1.92 \\
MiniLM-L6 v2 & Dense & 20.67 & 17.69 & 15.84 & 20.67 & 17.02 & \cellcolor{aliceblue}\textbf{14.57} & \cellcolor{aliceblue}\textbf{0.59} & 1.84 & \cellcolor{aliceblue}\textbf{6.24} & 0.00 & 0.48 & 1.44 \\
BGE-Large v1.5 & Dense & \cellcolor{aliceblue}\textbf{23.08} & \cellcolor{aliceblue}\textbf{19.26} & \cellcolor{aliceblue}\textbf{16.07} & \cellcolor{aliceblue}\textbf{23.08} & \cellcolor{aliceblue}\textbf{18.08} & 14.40 & 0.42 & \cellcolor{aliceblue}\textbf{1.94} & 6.12 & 0.00 & \cellcolor{aliceblue}\textbf{0.96} & \cellcolor{aliceblue}\textbf{3.37} \\
\hline
COLT ToolLens & Tool & 11.06 & 10.11 & 8.60 & 11.06 & 9.90 & 7.88 & 0.18 & 1.05 & 3.47 & 0.00 & 0.00 & 0.96 \\
COLT ToolBench & Tool & 12.50 & 12.64 & 11.57 & 12.50 & 12.60 & 10.84 & 0.28 & 1.49 & 4.53 & 0.00 & 0.48 & 1.92 \\
Tool-Embed & Tool & \cellcolor{aliceblue}\textbf{21.15} & \cellcolor{aliceblue}\textbf{20.08} & 17.19 & \cellcolor{aliceblue}\textbf{21.15} & \cellcolor{aliceblue}\textbf{19.81} & \cellcolor{aliceblue}\textbf{15.94} & \cellcolor{aliceblue}\textbf{0.42} & 2.01 & 6.41 & 0.00 & 0.48 & 1.92 \\
ToolRet & Tool & \cellcolor{aliceblue}\textbf{21.15} & 19.87 & \cellcolor{aliceblue}\textbf{17.21} & \cellcolor{aliceblue}\textbf{21.15} & 19.13 & 15.77 & 0.37 & \cellcolor{aliceblue}\textbf{2.02} & \cellcolor{aliceblue}\textbf{6.69} & 0.00 & \cellcolor{aliceblue}\textbf{0.96} & \cellcolor{aliceblue}\textbf{3.37} \\
\hline
E5-Mistral 7B & Decoder-only & 15.87 & 14.78 & 11.56 & 15.87 & 14.71 & 10.22 & 0.33 & 1.74 & 4.58 & 0.00 & \cellcolor{aliceblue}\textbf{0.48} & 1.92 \\
Qwen-Embedding 8B & Decoder-only & \cellcolor{aliceblue}\textbf{20.67} & \cellcolor{aliceblue}\textbf{18.84} & \cellcolor{aliceblue}\textbf{16.51} & \cellcolor{aliceblue}\textbf{20.67} & \cellcolor{aliceblue}\textbf{18.08} & \cellcolor{aliceblue}\textbf{15.29} & \cellcolor{aliceblue}\textbf{0.46} & \cellcolor{aliceblue}\textbf{1.87} & \cellcolor{aliceblue}\textbf{6.15} & 0.00 & 0.00 & \cellcolor{aliceblue}\textbf{2.40} \\
\bottomrule
\end{tabular}%
}
\caption{Retrieval results on Task Query $T_q$ (Single-Agent and Multi-Agent Task Queries) and Task Descriptions $T_d$. We \colorbox{aliceblue}{\textbf{highlight}} the best performance in each type of model.}
\label{tab:retrieval}
\end{table}

\begin{table}[!t]
\centering
\renewcommand{\arraystretch}{1.25}
\resizebox{\textwidth}{!}{%
\begin{tabular}{lccccccccccc}
\toprule
\textbf{Model} & \textbf{Type} & \multicolumn{3}{c}{\textbf{NDCG}} & \multicolumn{2}{c}{\textbf{Completeness}} & \multicolumn{3}{c}{\textbf{NDCG}} & \multicolumn{2}{c}{\textbf{Completeness}}\\
\cline{3-12}
 & & \textbf{@1} & \textbf{@5} & \textbf{@20} & \textbf{@1} & \textbf{@5} &\textbf{@1} & \textbf{@5} & \textbf{@20} & \textbf{@1} & \textbf{@5 } \\
\midrule
&&\multicolumn{5}{c}{\cellcolor{lavenderblush}\textit{Task Query}} &\multicolumn{5}{c}{\cellcolor{oldlace}\textit{Task Description}} \\
\hline
Random Shuffle & / & 51.43 & 50.98 & 74.61 & 28.50 & 61.00 & 53.00 & 48.27 & 76.60 & 0.00 & 0.00 \\
\hline
MiniLM-L12 v2 & Cross-Encoder & 48.27 & 48.06 & 73.53 & 28.50 & 52.00 & 48.00 & 57.49 & 81.58 & 0.00 & 0.00 \\
MXBAI Reranker Large & Cross-Encoder & 52.09 & 53.42 & 76.32 & 30.00 & 58.00 & 52.84 & \cellcolor{aliceblue}\textbf{61.33} & \cellcolor{aliceblue}\textbf{82.22} & 0.00 & 0.48 \\
BGE Reranker v2 & Cross-Encoder & \cellcolor{aliceblue}\textbf{63.09} & \cellcolor{aliceblue}\textbf{59.84} & \cellcolor{aliceblue}\textbf{79.59} & \cellcolor{aliceblue}\textbf{30.50} & \cellcolor{aliceblue}\textbf{59.00} & \cellcolor{aliceblue}\textbf{54.31} & 60.55 & 81.97 & 0.00 & \cellcolor{aliceblue}\textbf{0.96} \\
\hline
Tool-Rank 4B & Tool & 63.84 & \cellcolor{aliceblue}\textbf{64.45} & 81.78 & 28.00 & 59.50 & 52.24 & 61.12 & 82.32 & 0.00 & 0.48 \\
Tool-Rank 8B & Tool & \cellcolor{aliceblue}\textbf{66.67} & 64.36 & \cellcolor{aliceblue}\textbf{81.96} & \cellcolor{aliceblue}\textbf{30.50} & \cellcolor{aliceblue}\textbf{60.00} & \cellcolor{aliceblue}\textbf{53.92} & \cellcolor{aliceblue}\textbf{61.97} & \cellcolor{aliceblue}\textbf{82.76} & 0.00 & \cellcolor{aliceblue}\textbf{0.96} \\
\hline
MonoT5 Base MS-Marco & Decoder-only  & 57.84 & 57.37 & 78.69 & 27.00 & 60.00 & 52.31 & 59.10 & 81.33 & 0.00 & 0.96  \\
Qwen Reranker 0.6B & Decoder-only & 63.34 & 63.72 & 81.08 & \cellcolor{aliceblue}\textbf{30.50} & 58.50 & 53.96 & 60.47 & 82.04 & 0.00 & 0.48 \\
Qwen Reranker 4B & Decoder-only & \cellcolor{aliceblue}\textbf{64.67} & \cellcolor{aliceblue}\textbf{64.53} & \cellcolor{aliceblue}\textbf{81.97} & 29.50 & \cellcolor{aliceblue}\textbf{62.50} & \cellcolor{aliceblue}\textbf{58.00} & \cellcolor{aliceblue}\textbf{60.58} & \cellcolor{aliceblue}\textbf{82.84} & 0.00 & 0.00 \\
\hline
RankGPT Qwen-3 32B & LLM-based & 61.00 & 61.05 & 80.18 & 29.00 & \cellcolor{aliceblue}\textbf{58.50} & 53.95 & 60.78 & 82.36 & 0.00 & 0.48 \\
RankGPT LLaMA-3.3 70B & LLM-based & 61.09 & 59.92 & 79.80 & 29.50 & 53.50 & 52.60 & 61.24 & 82.34 & 0.00 & \cellcolor{aliceblue}\textbf{0.96} \\
RankGPT GPT-5.2 & LLM-based & \cellcolor{aliceblue}\textbf{63.59} & \cellcolor{aliceblue}\textbf{64.57} & \cellcolor{aliceblue}\textbf{81.88} & \cellcolor{aliceblue}\textbf{30.50} & 56.50 & \cellcolor{aliceblue}\textbf{66.00} & \cellcolor{aliceblue}\textbf{64.66} & \cellcolor{aliceblue}\textbf{84.69} & 0.00 & 0.00 \\
\bottomrule
\end{tabular}%
}
\caption{Reranking results on Task Query $T_q$ (Single-Agent and Multi-Agent Task Queries) and Task Descriptions $T_d$. We \colorbox{aliceblue}{\textbf{highlight}} the best performance in each type of model.}
\vspace{-0.5em}
\label{tab:rerank}
\end{table}

\subsection{Experimental Setup}

We evaluate agent search under both executable task queries and high-level task descriptions. For retrieval, methods search over the full agent repository using binary relevance labels derived from execution outcomes. For reranking, each method is given the top-20 agents with highest execution-based relevance and is evaluated against the golden ranking induced by aggregated subtask completion performance.

For retrieval, we report Precision, Recall, NDCG, and Completeness \citep{qu2024towards, shi2025retrieval}. For reranking, we report NDCG and Completeness using graded relevance labels. We compare representative methods from four retrieval families, including sparse, dense, tool-aware, and decoder-only embedding models \citep{robertson2009probabilistic, formal2021splade, santhanam2022colbertv2, izacard2021unsupervised, ni2022large, wang2021minilmv2, xiao2024cpack, shi2025retrieval, lu2025tools, qu2024towards}, and four reranking families, including cross-encoders, tool-specific rankers, decoder-only rerankers, and LLM-based rankers \citep{wang2021minilmv2, multi2024m3, li2025prorank, lu2025tools, nogueira2020document, qwen3embedding, sun2023chatgpt, ma2024fine}. All methods retrieve or rerank a fixed top-$K$ candidate set under the same evaluation protocol.

\subsection{Benchmarking Analysis}

\TAB~\ref{tab:retrieval} shows retrieval performance under execution-based relevance. On task queries, tool-aware retrievers outperform dense and sparse baselines, while on task descriptions, dense retrievers become more competitive, with BGE achieving the strongest overall performance. However, performance drops substantially when moving from executable queries to high-level task descriptions, and completeness remains low across all methods, highlighting the difficulty of retrieving agents that can fully satisfy abstract requirements. \textbf{Overall, these results indicate that while retrieval can capture coarse relevance, it struggles to identify agents with comprehensive task-solving capability, especially under high-level task specifications without explicit executable demands.}

\TAB~\ref{tab:rerank} shows reranking results on execution-grounded candidate pools. On task queries, different model families achieve broadly similar performance, suggesting that surface-level signals are often sufficient for relatively concrete tasks. On task descriptions, however, \textbf{decoder-only and LLM-based rerankers perform more strongly, possibly because their stronger generative capacity helps infer latent or implicit requirements behind high-level task descriptions.} Nevertheless, completeness remains limited, showing that improved ordering does not fully resolve the challenge of identifying agents that can completely satisfy complex requirements.

\begin{figure*}[t]
\centering
\begin{subfigure}[b]{0.30\textwidth}
\vspace{0pt}
\centering
\includegraphics[width=\linewidth]{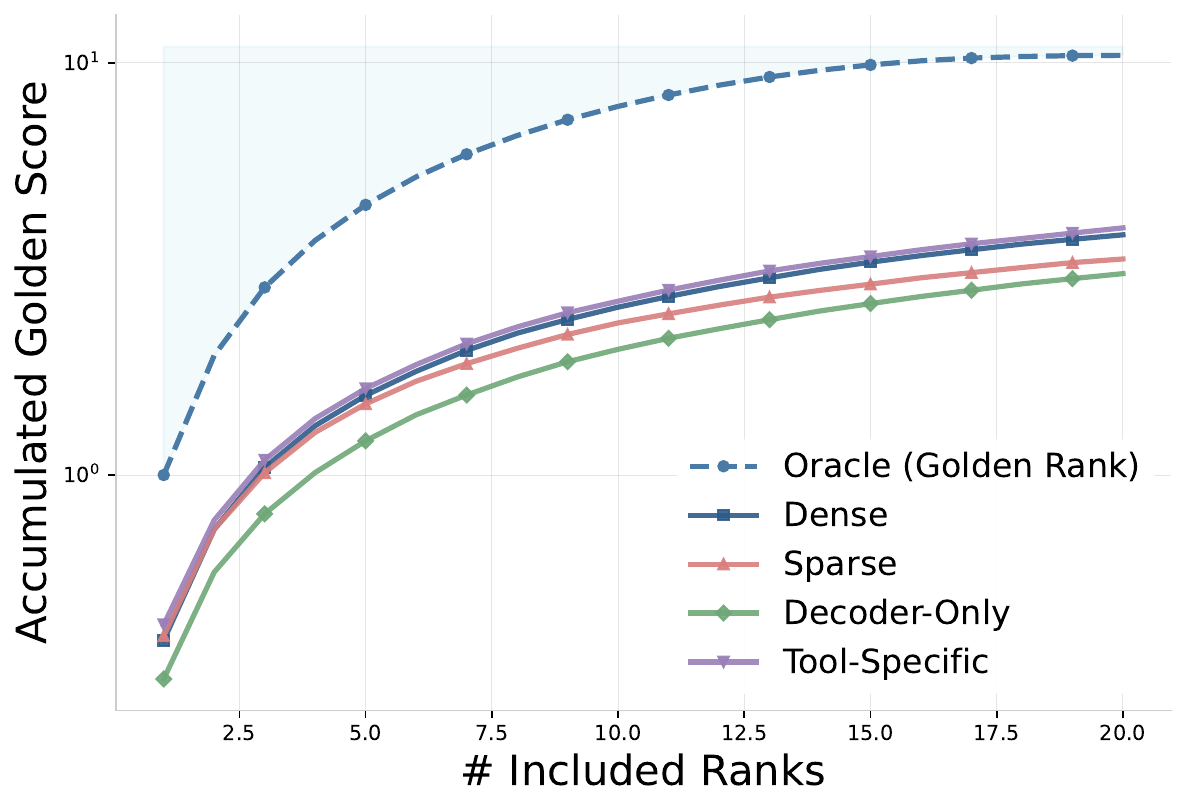}
\caption{Golden ranking accumulated agent scores on 2452 Single-Agent Task Queries. }
\label{fig:semantic-gap-single}
\end{subfigure}
\hfill
\begin{subfigure}[b]{0.30\textwidth}
\vspace{0pt}
\centering
\includegraphics[width=\linewidth]{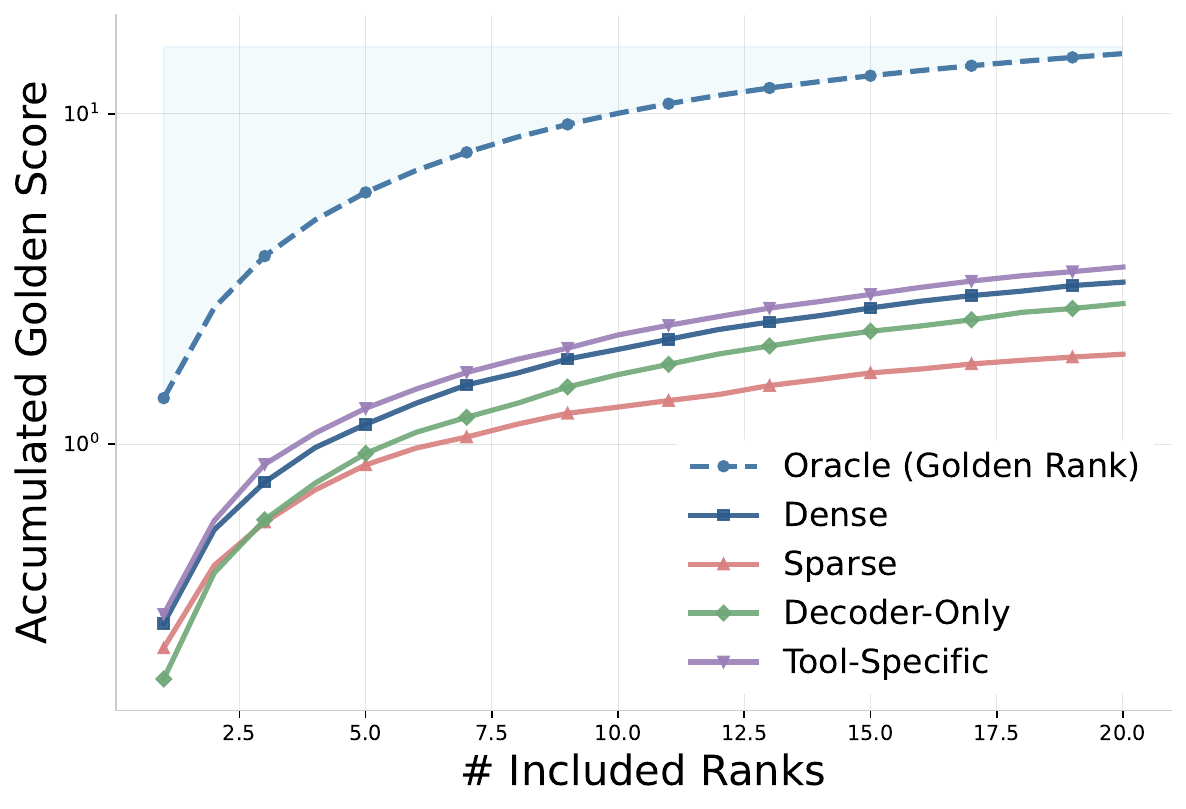}\caption{Golden ranking accumulated agent scores on 500 Multi-Agent Task Queries.}
\label{fig:semantic-gap-multi}
\end{subfigure}
\hfill
\begin{subfigure}[b]{0.30\textwidth}
\vspace{0pt}
\centering
\includegraphics[width=\linewidth]{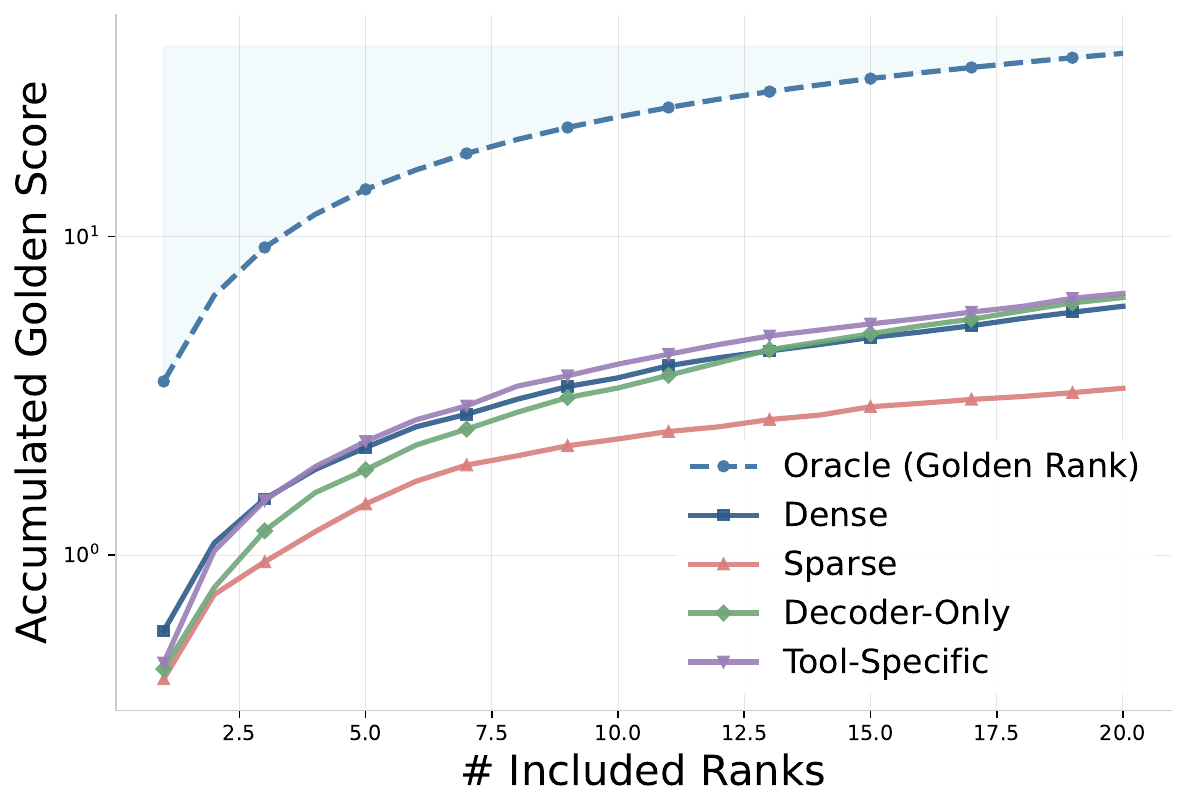}
\caption{Golden ranking accumulated agent scores on 259 Task Descriptions.}
\label{fig:semantic-gap-td}
\end{subfigure}
\caption{The Gap between Surface-matching and Execution.}
\label{fig:surface-gap}
\end{figure*}

To further examine this limitation, \FIG~\ref{fig:surface-gap} compares accumulated golden performance under model rankings with the oracle ranking. All methods remain substantially below the oracle, with gains distributed gradually rather than concentrated at top ranks. This indicates that many high-performing agents are ranked too low, reflecting \textbf{a misalignment between agent documentation and actual execution performance}. This misalignment leads to a persistent semantic–performance gap, where documentation-based matching fails to accurately capture execution effectiveness.

\subsection{Benchmark Validation}

\begin{figure*}[t]
\centering

\begin{subfigure}[b]{0.36\textwidth}
\vspace{0pt}
\centering
\includegraphics[width=\linewidth]{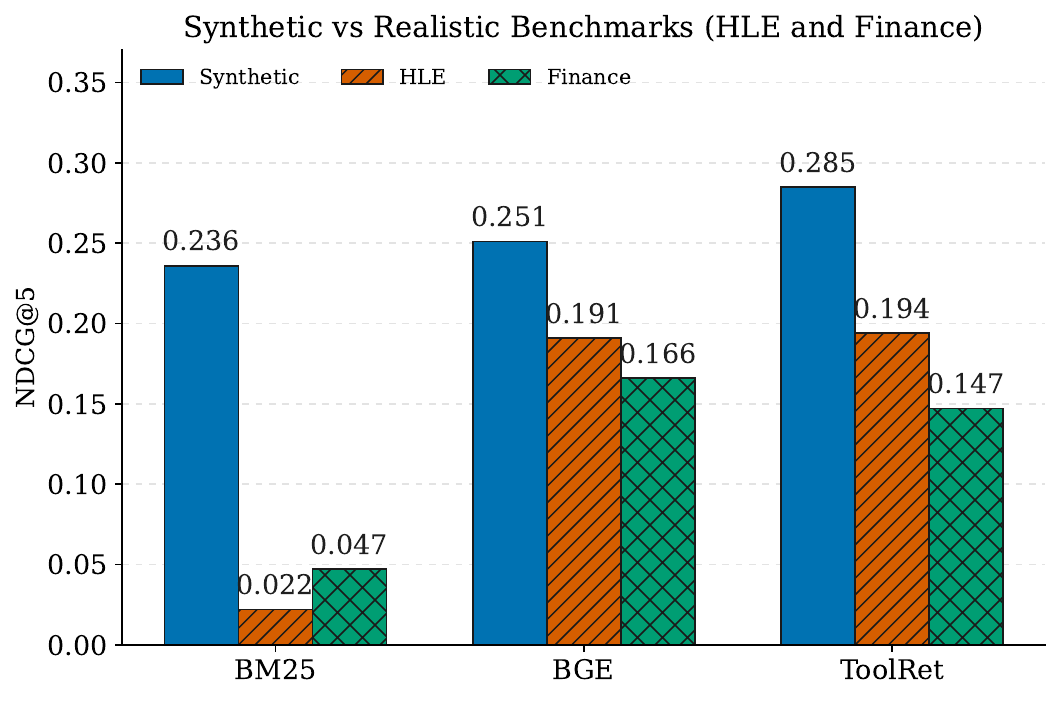}
\caption{NDCG@5: Comparison between Realistic and Synthetic Single-agent Task Queries.}
\label{fig:synthetic_realistic}
\end{subfigure}
\hfill
\begin{subfigure}[b]{0.3\textwidth}
\vspace{0pt}
\centering
\includegraphics[width=\linewidth]{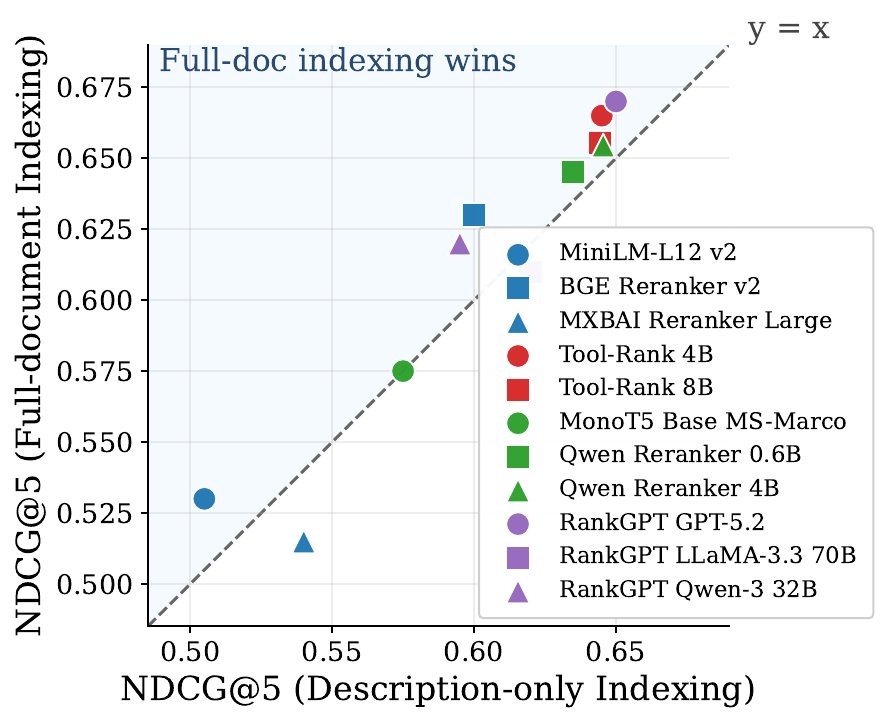}
\caption{NDCG@5: description-only vs full-document indexing on 2,952 Task Queries.}
\label{fig:indexing_retrieval}
\end{subfigure}
\hfill
\begin{subfigure}[b]{0.3\textwidth}
\vspace{0pt}
\centering
\includegraphics[width=\linewidth]{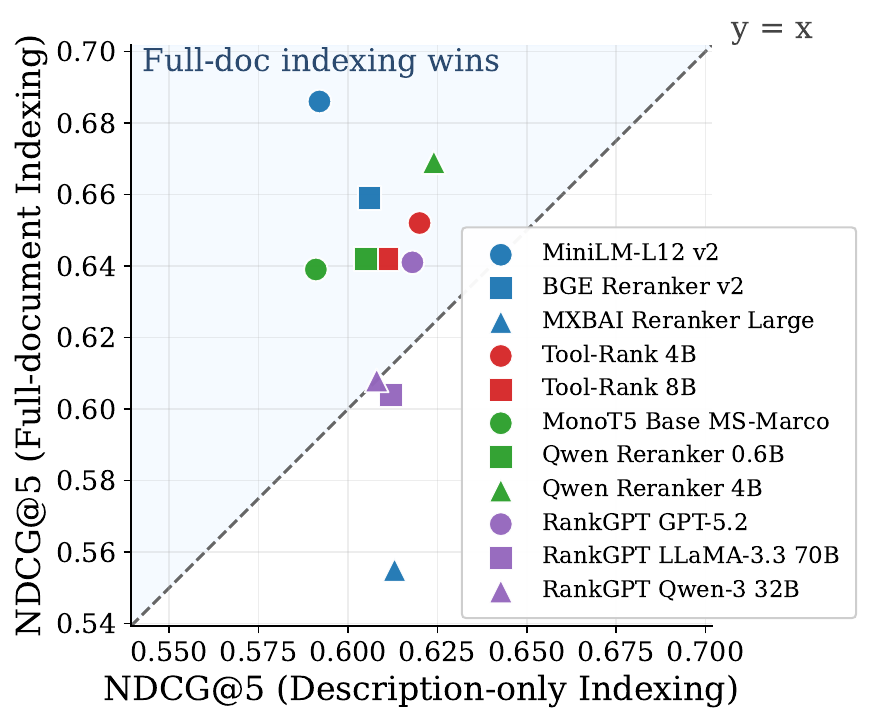}
\caption{NDCG@5: description-only vs full-document indexing on 259 Task Descriptions.}
\label{fig:indexing_reranking}
\end{subfigure}

\caption{
Comparison between indexing and query realism.}
\label{fig:rq6_rq4_analysis}
\end{figure*}

We validate the realism of our benchmark by comparing retrieval trends on synthetic queries with those on external realistic benchmarks. As shown in Figure~\ref{fig:synthetic_realistic}, representative sparse, dense, and tool-aware retrievers exhibit consistent relative trends across settings: dense and tool-aware methods remain substantially stronger than sparse retrieval, while absolute performance on realistic queries is notably lower. \textbf{This suggests that our synthetic benchmark preserves the relative performance ordering of different methods while providing a controlled evaluation setting.} In contrast, realistic queries are inherently more difficult, as they do not guarantee the existence of highly relevant agents in the candidate pool, resulting in lower absolute performance.

We further validate the reliability of LLM-based relevance annotation by comparing it with human judgments. Following the relevance label in our \OURSBENCH{}, we focus on binary relevance signals by grouping scores of 4–5 as positive and 1–3 as negative, which are the primary signals used in our benchmark. We conduct a human evaluation on 500 execution instances with three AI PhD-level annotators and observe high agreement between LLM-based and human judgments, with a Cohen's kappa of $\kappa=0.93$ and accuracy $96.67\%$. \textbf{These results indicate that LLM-as-judge provides reliable supervision for large-scale evaluation, supporting its use for constructing execution-grounded relevance labels.}

\subsection{Execution-Aware Probing}

We next study whether lightweight behavioral signals can improve agent ranking. First, we compare description-only indexing with full-document indexing that additionally includes usage examples. As shown in \FIG~\ref{fig:indexing_retrieval} and \FIG~\ref{fig:indexing_reranking}, most methods improve under full-document indexing. \textbf{These observations suggest that usage examples, often provided and verified by developers through execution, offer behavioral evidence beyond static descriptions.}

\begin{figure*}[t]
\centering
\begin{subfigure}[b]{0.5\textwidth}
\vspace{0pt}
\centering
\includegraphics[width=\textwidth]{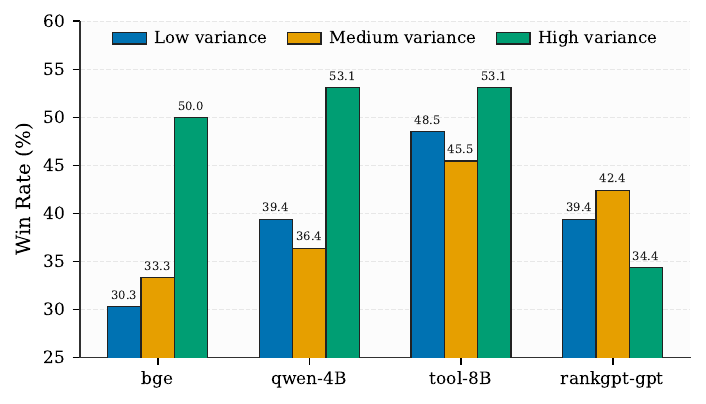}
\caption{NDCG@5: win rate v.s. probe score variance}
\label{fig:winrate}
\end{subfigure}
\hspace{1em}
\begin{subfigure}[b]{0.4\textwidth}
\vspace{0pt}
\centering
\renewcommand{\arraystretch}{1.25}
\resizebox{\textwidth}{!}{%
\begin{tabular}{lll}
\toprule
\textbf{Model} & {\textbf{NDCG@5}} & {\textbf{NDCG@10}}\\
\midrule
BGE Reranker v2            & 57.93              & 80.76              \\
\quad w/ Probing           & 58.16~\up{+0.40\%} & 80.82~\up{+0.07\%} \\
\hline
Tool-Rank 8B               & 60.82              & 82.09              \\
\quad w/ Probing           & 61.71~\up{+1.46\%} & 82.24~\up{+0.18\%} \\
\hline
Qwen Reranker 4B           & 60.96              & 82.12              \\
\quad w/ Probing           & 61.91~\up{+1.56\%} & 82.38~\up{+0.32\%} \\
\hline
RankGPT GPT-5.2            & 61.25              & 82.40              \\
\quad w/ Probing           & 59.60~\dn{-2.69\%} & 82.94~\up{+0.66\%} \\
\bottomrule
\end{tabular}
}
\caption{Execution-Aware Probing enhancements on 100 Task Descriptions $T_d$.}
\label{tab:rerank_probing}
\end{subfigure}
\caption{Execution-Aware Probing for Reranking on Task Description.}
\vspace{-1em}
\label{fig:probing}
\end{figure*}

We then investigate explicit probing, where LLMs generate probing queries, candidate agents are executed on these queries, and the resulting responses are used as additional ranking signals. \FIG~\ref{fig:winrate} shows that probing is most effective when the probing responses exhibit medium or high variance across agents, whereas low-variance probes provide limited discrimination. Using these execution-derived signals, most rerankers achieve consistent improvements as shown in \FIG~\ref{tab:rerank_probing}, demonstrating that \textbf{lightweight behavioral probing can complement description-based ranking and better capture execution-level capability}.

\section{Conclusion}

We introduce \OURSBENCH{}, a large-scale benchmark for agent search in open ecosystems, covering nearly 10,000 real-world agents and supporting both executable task queries and high-level task descriptions. By grounding relevance in execution outcomes, our benchmark reveals a substantial semantic–performance gap: methods based on textual similarity often fail to identify the best-performing agents. While existing retrieval and reranking approaches provide useful coarse signals, they remain limited in capturing execution-dependent capability, especially for abstract and multi-step tasks. We show that incorporating lightweight behavioral signals, such as richer indexing and execution-aware probing, can improve ranking quality. These results highlight the importance of execution-aware methods and establish \OURSBENCH{} as a practical testbed for performance-grounded agent search.




\bibliography{main}

@article{robertson2009probabilistic,
  author       = {Stephen E. Robertson and
                  Hugo Zaragoza},
  title        = {The Probabilistic Relevance Framework: {BM25} and Beyond},
  journal      = {Found. Trends Inf. Retr.},
  volume       = {3},
  number       = {4},
  pages        = {333--389},
  year         = {2009},
  url          = {https://doi.org/10.1561/1500000019},
  doi          = {10.1561/1500000019},
  bibsource    = {dblp computer science bibliography, https://dblp.org}
}

@article{formal2021splade,
  author       = {Thibault Formal and
                  Carlos Lassance and
                  Benjamin Piwowarski and
                  St{\'{e}}phane Clinchant},
  title        = {{SPLADE} v2: Sparse Lexical and Expansion Model for Information Retrieval},
  journal      = {CoRR},
  volume       = {abs/2109.10086},
  year         = {2021},
  url          = {https://arxiv.org/abs/2109.10086},
  eprinttype    = {arXiv},
  eprint       = {2109.10086},
  bibsource    = {dblp computer science bibliography, https://dblp.org}
}

@inproceedings{santhanam2022colbertv2,
  author       = {Keshav Santhanam and
                  Omar Khattab and
                  Jon Saad{-}Falcon and
                  Christopher Potts and
                  Matei Zaharia},
  editor       = {Marine Carpuat and
                  Marie{-}Catherine de Marneffe and
                  Iv{\'{a}}n Vladimir Meza Ru{\'{\i}}z},
  title        = {ColBERTv2: Effective and Efficient Retrieval via Lightweight Late
                  Interaction},
  booktitle    = {Proceedings of the 2022 Conference of the North American Chapter of
                  the Association for Computational Linguistics: Human Language Technologies,
                  {NAACL} 2022, Seattle, WA, United States, July 10-15, 2022},
  pages        = {3715--3734},
  publisher    = {Association for Computational Linguistics},
  year         = {2022},
  url          = {https://doi.org/10.18653/v1/2022.naacl-main.272},
  doi          = {10.18653/V1/2022.NAACL-MAIN.272},
  bibsource    = {dblp computer science bibliography, https://dblp.org}
}

@inproceedings{xiao2024cpack,
  author       = {Shitao Xiao and
                  Zheng Liu and
                  Peitian Zhang and
                  Niklas Muennighoff and
                  Defu Lian and
                  Jian{-}Yun Nie},
  editor       = {Grace Hui Yang and
                  Hongning Wang and
                  Sam Han and
                  Claudia Hauff and
                  Guido Zuccon and
                  Yi Zhang},
  title        = {C-Pack: Packed Resources For General Chinese Embeddings},
  booktitle    = {Proceedings of the 47th International {ACM} {SIGIR} Conference on
                  Research and Development in Information Retrieval, {SIGIR} 2024, Washington
                  DC, USA, July 14-18, 2024},
  pages        = {641--649},
  publisher    = {{ACM}},
  year         = {2024},
  url          = {https://doi.org/10.1145/3626772.3657878},
  doi          = {10.1145/3626772.3657878},
  bibsource    = {dblp computer science bibliography, https://dblp.org}
}

@inproceedings{shi2025retrieval,
  author       = {Zhengliang Shi and
                  Yuhan Wang and
                  Lingyong Yan and
                  Pengjie Ren and
                  Shuaiqiang Wang and
                  Dawei Yin and
                  Zhaochun Ren},
  editor       = {Wanxiang Che and
                  Joyce Nabende and
                  Ekaterina Shutova and
                  Mohammad Taher Pilehvar},
  title        = {Retrieval Models Aren't Tool-Savvy: Benchmarking Tool Retrieval for
                  Large Language Models},
  booktitle    = {Findings of the Association for Computational Linguistics, {ACL} 2025,
                  Vienna, Austria, July 27 - August 1, 2025},
  series       = {Findings of {ACL}},
  volume       = {{ACL} 2025},
  pages        = {24497--24524},
  publisher    = {Association for Computational Linguistics},
  year         = {2025},
  url          = {https://aclanthology.org/2025.findings-acl.1258/},
  bibsource    = {dblp computer science bibliography, https://dblp.org}
}

@article{behnamghader2024llm2vec,
  author       = {Parishad BehnamGhader and
                  Vaibhav Adlakha and
                  Marius Mosbach and
                  Dzmitry Bahdanau and
                  Nicolas Chapados and
                  Siva Reddy},
  title        = {LLM2Vec: Large Language Models Are Secretly Powerful Text Encoders},
  journal      = {CoRR},
  volume       = {abs/2404.05961},
  year         = {2024},
  url          = {https://doi.org/10.48550/arXiv.2404.05961},
  doi          = {10.48550/ARXIV.2404.05961},
  eprinttype    = {arXiv},
  eprint       = {2404.05961},
  bibsource    = {dblp computer science bibliography, https://dblp.org}
}

@article{lu2025tools,
  author       = {Xuan Lu and
                  Haohang Huang and
                  Rui Meng and
                  Yaohui Jin and
                  Wenjun Zeng and
                  Xiaoyu Shen},
  title        = {Tools are under-documented: Simple Document Expansion Boosts Tool
                  Retrieval},
  journal      = {CoRR},
  volume       = {abs/2510.22670},
  year         = {2025},
  url          = {https://doi.org/10.48550/arXiv.2510.22670},
  doi          = {10.48550/ARXIV.2510.22670},
  eprinttype    = {arXiv},
  eprint       = {2510.22670},
  bibsource    = {dblp computer science bibliography, https://dblp.org}
}

@inproceedings{wang2024improving,
  author       = {Liang Wang and
                  Nan Yang and
                  Xiaolong Huang and
                  Linjun Yang and
                  Rangan Majumder and
                  Furu Wei},
  editor       = {Lun{-}Wei Ku and
                  Andre Martins and
                  Vivek Srikumar},
  title        = {Improving Text Embeddings with Large Language Models},
  booktitle    = {Proceedings of the 62nd Annual Meeting of the Association for Computational
                  Linguistics (Volume 1: Long Papers), {ACL} 2024, Bangkok, Thailand,
                  August 11-16, 2024},
  pages        = {11897--11916},
  publisher    = {Association for Computational Linguistics},
  year         = {2024},
  url          = {https://doi.org/10.18653/v1/2024.acl-long.642},
  doi          = {10.18653/V1/2024.ACL-LONG.642},
  bibsource    = {dblp computer science bibliography, https://dblp.org}
}

@article{izacard2021unsupervised,
  author       = {Gautier Izacard and
                  Mathilde Caron and
                  Lucas Hosseini and
                  Sebastian Riedel and
                  Piotr Bojanowski and
                  Armand Joulin and
                  Edouard Grave},
  title        = {Unsupervised Dense Information Retrieval with Contrastive Learning},
  journal      = {Trans. Mach. Learn. Res.},
  volume       = {2022},
  year         = {2022},
  url          = {https://openreview.net/forum?id=jKN1pXi7b0},
  bibsource    = {dblp computer science bibliography, https://dblp.org}
}

@inproceedings{ni2022large,
  author       = {Jianmo Ni and
                  Chen Qu and
                  Jing Lu and
                  Zhuyun Dai and
                  Gustavo Hern{\'{a}}ndez {\'{A}}brego and
                  Ji Ma and
                  Vincent Y. Zhao and
                  Yi Luan and
                  Keith B. Hall and
                  Ming{-}Wei Chang and
                  Yinfei Yang},
  editor       = {Yoav Goldberg and
                  Zornitsa Kozareva and
                  Yue Zhang},
  title        = {Large Dual Encoders Are Generalizable Retrievers},
  booktitle    = {Proceedings of the 2022 Conference on Empirical Methods in Natural
                  Language Processing, {EMNLP} 2022, Abu Dhabi, United Arab Emirates,
                  December 7-11, 2022},
  pages        = {9844--9855},
  publisher    = {Association for Computational Linguistics},
  year         = {2022},
  url          = {https://doi.org/10.18653/v1/2022.emnlp-main.669},
  doi          = {10.18653/V1/2022.EMNLP-MAIN.669},
  bibsource    = {dblp computer science bibliography, https://dblp.org}
}

@inproceedings{wang2021minilmv2,
  author       = {Wenhui Wang and
                  Hangbo Bao and
                  Shaohan Huang and
                  Li Dong and
                  Furu Wei},
  editor       = {Chengqing Zong and
                  Fei Xia and
                  Wenjie Li and
                  Roberto Navigli},
  title        = {MiniLMv2: Multi-Head Self-Attention Relation Distillation for Compressing
                  Pretrained Transformers},
  booktitle    = {Findings of the Association for Computational Linguistics: {ACL/IJCNLP}
                  2021, Online Event, August 1-6, 2021},
  series       = {Findings of {ACL}},
  volume       = {{ACL-IJCNLP} 2021},
  pages        = {2140--2151},
  publisher    = {Association for Computational Linguistics},
  year         = {2021},
  url          = {https://doi.org/10.18653/v1/2021.findings-acl.188},
  doi          = {10.18653/V1/2021.FINDINGS-ACL.188},
  bibsource    = {dblp computer science bibliography, https://dblp.org}
}

@online{rerank2024mxbai,
  title={Boost Your Search With The Crispy Mixedbread Rerank Models},
  author={Aamir Shakir and Darius Koenig and Julius Lipp and Sean Lee},
  year={2024},
  url={https://www.mixedbread.com/blog/mxbai-rerank-v1},
}

@inproceedings{qu2024towards,
  author       = {Changle Qu and
                  Sunhao Dai and
                  Xiaochi Wei and
                  Hengyi Cai and
                  Shuaiqiang Wang and
                  Dawei Yin and
                  Jun Xu and
                  Ji{-}Rong Wen},
  editor       = {Edoardo Serra and
                  Francesca Spezzano},
  title        = {Towards Completeness-Oriented Tool Retrieval for Large Language Models},
  booktitle    = {Proceedings of the 33rd {ACM} International Conference on Information
                  and Knowledge Management, {CIKM} 2024, Boise, ID, USA, October 21-25,
                  2024},
  pages        = {1930--1940},
  publisher    = {{ACM}},
  year         = {2024},
  url          = {https://doi.org/10.1145/3627673.3679847},
  doi          = {10.1145/3627673.3679847},
  bibsource    = {dblp computer science bibliography, https://dblp.org}
}

@inproceedings{nogueira2020document,
  author       = {Rodrigo Nogueira and
                  Zhiying Jiang and
                  Ronak Pradeep and
                  Jimmy Lin},
  editor       = {Trevor Cohn and
                  Yulan He and
                  Yang Liu},
  title        = {Document Ranking with a Pretrained Sequence-to-Sequence Model},
  booktitle    = {Findings of the Association for Computational Linguistics: {EMNLP}
                  2020, Online Event, 16-20 November 2020},
  series       = {Findings of {ACL}},
  volume       = {{EMNLP} 2020},
  pages        = {708--718},
  publisher    = {Association for Computational Linguistics},
  year         = {2020},
  url          = {https://doi.org/10.18653/v1/2020.findings-emnlp.63},
  doi          = {10.18653/V1/2020.FINDINGS-EMNLP.63},
  bibsource    = {dblp computer science bibliography, https://dblp.org}
}

@inproceedings{multi2024m3,
  author       = {Jianlyu Chen and
                  Shitao Xiao and
                  Peitian Zhang and
                  Kun Luo and
                  Defu Lian and
                  Zheng Liu},
  editor       = {Lun{-}Wei Ku and
                  Andre Martins and
                  Vivek Srikumar},
  title        = {M3-Embedding: Multi-Linguality, Multi-Functionality, Multi-Granularity
                  Text Embeddings Through Self-Knowledge Distillation},
  booktitle    = {Findings of the Association for Computational Linguistics, {ACL} 2024,
                  Bangkok, Thailand and virtual meeting, August 11-16, 2024},
  series       = {Findings of {ACL}},
  volume       = {{ACL} 2024},
  pages        = {2318--2335},
  publisher    = {Association for Computational Linguistics},
  year         = {2024},
  url          = {https://doi.org/10.18653/v1/2024.findings-acl.137},
  doi          = {10.18653/V1/2024.FINDINGS-ACL.137},
  bibsource    = {dblp computer science bibliography, https://dblp.org}
}

@inproceedings{zheng2403toolrerank,
  author       = {Yuanhang Zheng and
                  Peng Li and
                  Wei Liu and
                  Yang Liu and
                  Jian Luan and
                  Bin Wang},
  editor       = {Nicoletta Calzolari and
                  Min{-}Yen Kan and
                  V{\'{e}}ronique Hoste and
                  Alessandro Lenci and
                  Sakriani Sakti and
                  Nianwen Xue},
  title        = {ToolRerank: Adaptive and Hierarchy-Aware Reranking for Tool Retrieval},
  booktitle    = {Proceedings of the 2024 Joint International Conference on Computational
                  Linguistics, Language Resources and Evaluation, {LREC/COLING} 2024,
                  20-25 May, 2024, Torino, Italy},
  pages        = {16263--16273},
  publisher    = {{ELRA} and {ICCL}},
  year         = {2024},
  url          = {https://aclanthology.org/2024.lrec-main.1413},
  bibsource    = {dblp computer science bibliography, https://dblp.org}
}

@inproceedings{sun2023chatgpt,
  author       = {Weiwei Sun and
                  Lingyong Yan and
                  Xinyu Ma and
                  Shuaiqiang Wang and
                  Pengjie Ren and
                  Zhumin Chen and
                  Dawei Yin and
                  Zhaochun Ren},
  editor       = {Houda Bouamor and
                  Juan Pino and
                  Kalika Bali},
  title        = {Is ChatGPT Good at Search? Investigating Large Language Models as
                  Re-Ranking Agents},
  booktitle    = {Proceedings of the 2023 Conference on Empirical Methods in Natural
                  Language Processing, {EMNLP} 2023, Singapore, December 6-10, 2023},
  pages        = {14918--14937},
  publisher    = {Association for Computational Linguistics},
  year         = {2023},
  url          = {https://doi.org/10.18653/v1/2023.emnlp-main.923},
  doi          = {10.18653/V1/2023.EMNLP-MAIN.923},
  bibsource    = {dblp computer science bibliography, https://dblp.org}
}

@inproceedings{ma2024fine,
  author       = {Xueguang Ma and
                  Liang Wang and
                  Nan Yang and
                  Furu Wei and
                  Jimmy Lin},
  editor       = {Grace Hui Yang and
                  Hongning Wang and
                  Sam Han and
                  Claudia Hauff and
                  Guido Zuccon and
                  Yi Zhang},
  title        = {Fine-Tuning LLaMA for Multi-Stage Text Retrieval},
  booktitle    = {Proceedings of the 47th International {ACM} {SIGIR} Conference on
                  Research and Development in Information Retrieval, {SIGIR} 2024, Washington
                  DC, USA, July 14-18, 2024},
  pages        = {2421--2425},
  publisher    = {{ACM}},
  year         = {2024},
  url          = {https://doi.org/10.1145/3626772.3657951},
  doi          = {10.1145/3626772.3657951},
  bibsource    = {dblp computer science bibliography, https://dblp.org}
}

@inproceedings{
sharma2026researchrubrics,
title={ResearchRubrics: A Benchmark of Prompts and Rubrics For Deep Research Agents},
author={Manasi Sharma and Chen Bo Calvin Zhang and Chaithanya Bandi and Ankit Aich and Huy Nghiem and Tahseen Rabbani and Ye Htet and Brian Jang and Aishwarya Balwani and Sumana Basu and Denis Peskoff and Clinton Wang and Marcos Ayestaran and Sean M. Hendryx and Brad Kenstler and Bing Liu},
booktitle={The Fourteenth International Conference on Learning Representations},
year={2026},
url={https://openreview.net/pdf?id=ErnvfmSX0P},
}

@inproceedings{DBLP:conf/iclr/ShangLZMLXL25,
  author       = {Yu Shang and
                  Yu Li and
                  Keyu Zhao and
                  Likai Ma and
                  Jiahe Liu and
                  Fengli Xu and
                  Yong Li},
  title        = {AgentSquare: Automatic {LLM} Agent Search in Modular Design Space},
  booktitle    = {The Thirteenth International Conference on Learning Representations,
                  {ICLR} 2025, Singapore, April 24-28, 2025},
  publisher    = {OpenReview.net},
  year         = {2025},
  url          = {https://openreview.net/forum?id=mPdmDYIQ7f},
  bibsource    = {dblp computer science bibliography, https://dblp.org}
}

@inproceedings{DBLP:conf/iclr/QinLYZYLLCTQZHT24,
  author       = {Yujia Qin and
                  Shihao Liang and
                  Yining Ye and
                  Kunlun Zhu and
                  Lan Yan and
                  Yaxi Lu and
                  Yankai Lin and
                  Xin Cong and
                  Xiangru Tang and
                  Bill Qian and
                  Sihan Zhao and
                  Lauren Hong and
                  Runchu Tian and
                  Ruobing Xie and
                  Jie Zhou and
                  Mark Gerstein and
                  Dahai Li and
                  Zhiyuan Liu and
                  Maosong Sun},
  title        = {ToolLLM: Facilitating Large Language Models to Master 16000+ Real-world
                  APIs},
  booktitle    = {The Twelfth International Conference on Learning Representations,
                  {ICLR} 2024, Vienna, Austria, May 7-11, 2024},
  publisher    = {OpenReview.net},
  year         = {2024},
  url          = {https://openreview.net/forum?id=dHng2O0Jjr},
  bibsource    = {dblp computer science bibliography, https://dblp.org}
}

@inproceedings{
yuan2025automated,
title={Automated Composition of Agents: A Knapsack Approach for Agentic Component Selection},
author={Michelle Yuan and Khushbu Pahwa and Shuaichen Chang and Mustafa Devrim Kaba and MONICA SUNKARA and Jiarong Jiang and Xiaofei Ma and Yi Zhang},
booktitle={The Thirty-ninth Annual Conference on Neural Information Processing Systems},
year={2025},
url={https://openreview.net/forum?id=1LPPMAUlaT}
}

@article{DBLP:journals/corr/abs-2508-07407,
  author       = {Jinyuan Fang and
                  Yanwen Peng and
                  Xi Zhang and
                  Yingxu Wang and
                  Xinhao Yi and
                  Guibin Zhang and
                  Yi Xu and
                  Bin Wu and
                  Siwei Liu and
                  Zihao Li and
                  Zhaochun Ren and
                  Nikos Aletras and
                  Xi Wang and
                  Han Zhou and
                  Zaiqiao Meng},
  title        = {A Comprehensive Survey of Self-Evolving {AI} Agents: {A} New Paradigm
                  Bridging Foundation Models and Lifelong Agentic Systems},
  journal      = {CoRR},
  volume       = {abs/2508.07407},
  year         = {2025},
  url          = {https://doi.org/10.48550/arXiv.2508.07407},
  doi          = {10.48550/ARXIV.2508.07407},
  eprinttype   = {arXiv},
  eprint       = {2508.07407},
  bibsource    = {dblp computer science bibliography, https://dblp.org}
}

@article{DBLP:journals/tmlr/GaoGHHJLLQQRWWXZZZXFZLQ26,
  author       = {Huan{-}ang Gao and
                  Jiayi Geng and
                  Wenyue Hua and
                  Mengkang Hu and
                  Xinzhe Juan and
                  Hongzhang Liu and
                  Shilong Liu and
                  Jiahao Qiu and
                  Xuan Qi and
                  Qihan Ren and
                  Yiran Wu and
                  Hongru Wang and
                  Han Xiao and
                  Yuhang Zhou and
                  Shaokun Zhang and
                  Jiayi Zhang and
                  Jinyu Xiang and
                  Yixiong Fang and
                  Qiwen Zhao and
                  Dongrui Liu and
                  Cheng Qian and
                  Zhenhailong Wang and
                  Minda Hu and
                  Huazheng Wang and
                  Qingyun Wu and
                  Heng Ji and
                  Mengdi Wang},
  title        = {A Survey of Self-Evolving Agents: What, When, How, and Where to Evolve
                  on the Path to Artificial Super Intelligence},
  journal      = {Trans. Mach. Learn. Res.},
  volume       = {2026},
  year         = {2026},
  url          = {https://openreview.net/forum?id=CTr3bovS5F},
  bibsource    = {dblp computer science bibliography, https://dblp.org}
}

@article{DBLP:journals/corr/abs-2504-19678,
  author       = {Mohamed Amine Ferrag and
                  Norbert Tihanyi and
                  M{\'{e}}rouane Debbah},
  title        = {From {LLM} Reasoning to Autonomous {AI} Agents: {A} Comprehensive
                  Review},
  journal      = {CoRR},
  volume       = {abs/2504.19678},
  year         = {2025},
  url          = {https://doi.org/10.48550/arXiv.2504.19678},
  doi          = {10.48550/ARXIV.2504.19678},
  eprinttype   = {arXiv},
  eprint       = {2504.19678},
  bibsource    = {dblp computer science bibliography, https://dblp.org}
}

@article{DBLP:journals/corr/abs-2402-02716,
  author       = {Xu Huang and
                  Weiwen Liu and
                  Xiaolong Chen and
                  Xingmei Wang and
                  Hao Wang and
                  Defu Lian and
                  Yasheng Wang and
                  Ruiming Tang and
                  Enhong Chen},
  title        = {Understanding the planning of {LLM} agents: {A} survey},
  journal      = {CoRR},
  volume       = {abs/2402.02716},
  year         = {2024},
  url          = {https://doi.org/10.48550/arXiv.2402.02716},
  doi          = {10.48550/ARXIV.2402.02716},
  eprinttype   = {arXiv},
  eprint       = {2402.02716},
  bibsource    = {dblp computer science bibliography, https://dblp.org}
}

@article{DBLP:journals/csur/QinHLCDCZZHXHFSWQTZLSXZ25,
  author       = {Yujia Qin and
                  Shengding Hu and
                  Yankai Lin and
                  Weize Chen and
                  Ning Ding and
                  Ganqu Cui and
                  Zheni Zeng and
                  Xuanhe Zhou and
                  Yufei Huang and
                  Chaojun Xiao and
                  Chi Han and
                  Yi R. Fung and
                  Yusheng Su and
                  Huadong Wang and
                  Cheng Qian and
                  Runchu Tian and
                  Kunlun Zhu and
                  Shihao Liang and
                  Xingyu Shen and
                  Bokai Xu and
                  Zhen Zhang and
                  Yining Ye and
                  Bowen Li and
                  Ziwei Tang and
                  Jing Yi and
                  Yuzhang Zhu and
                  Zhenning Dai and
                  Lan Yan and
                  Xin Cong and
                  Yaxi Lu and
                  Weilin Zhao and
                  Yuxiang Huang and
                  Junxi Yan and
                  Xu Han and
                  Xian Sun and
                  Dahai Li and
                  Jason Phang and
                  Cheng Yang and
                  Tongshuang Wu and
                  Heng Ji and
                  Guoliang Li and
                  Zhiyuan Liu and
                  Maosong Sun},
  title        = {Tool Learning with Foundation Models},
  journal      = {{ACM} Comput. Surv.},
  volume       = {57},
  number       = {4},
  pages        = {101:1--101:40},
  year         = {2025},
  url          = {https://doi.org/10.1145/3704435},
  doi          = {10.1145/3704435},
  bibsource    = {dblp computer science bibliography, https://dblp.org}
}

@inproceedings{
hu2025owl,
title={{OWL}: Optimized Workforce Learning for General Multi-Agent Assistance in Real-World Task Automation},
author={Mengkang Hu and Yuhang Zhou and Wendong Fan and Yuzhou Nie and Ziyu Ye and Bowei Xia and Tao Sun and Zhaoxuan Jin and Yingru Li and Zeyu Zhang and Yifeng Wang and Qianshuo Ye and Bernard Ghanem and Ping Luo and Guohao Li},
booktitle={The Thirty-ninth Annual Conference on Neural Information Processing Systems},
year={2025},
url={https://openreview.net/forum?id=MBJ46gd1CT}
}

@article{DBLP:journals/corr/abs-2411-04468,
  author       = {Adam Fourney and
                  Gagan Bansal and
                  Hussein Mozannar and
                  Cheng Tan and
                  Eduardo Salinas and
                  Erkang Zhu and
                  Friederike Niedtner and
                  Grace Proebsting and
                  Griffin Bassman and
                  Jack Gerrits and
                  Jacob Alber and
                  Peter Chang and
                  Ricky Loynd and
                  Robert West and
                  Victor Dibia and
                  Ahmed Awadallah and
                  Ece Kamar and
                  Rafah Hosn and
                  Saleema Amershi},
  title        = {Magentic-One: {A} Generalist Multi-Agent System for Solving Complex
                  Tasks},
  journal      = {CoRR},
  volume       = {abs/2411.04468},
  year         = {2024},
  url          = {https://doi.org/10.48550/arXiv.2411.04468},
  doi          = {10.48550/ARXIV.2411.04468},
  eprinttype   = {arXiv},
  eprint       = {2411.04468},
  bibsource    = {dblp computer science bibliography, https://dblp.org}
}

@inproceedings{wu2025joint,
  title={A joint optimization framework for enhancing efficiency of tool utilization in llm agents},
  author={Wu, Bin and Meij, Edgar and Yilmaz, Emine},
  booktitle={Findings of the Association for Computational Linguistics: ACL 2025},
  pages={22361--22373},
  year={2025}
}

@article{DBLP:journals/corr/abs-2601-15621,
  author       = {Qwen Team},
  title        = {Qwen3-TTS Technical Report},
  journal      = {CoRR},
  volume       = {abs/2601.15621},
  year         = {2026},
  url          = {https://doi.org/10.48550/arXiv.2601.15621},
  doi          = {10.48550/ARXIV.2601.15621},
  eprinttype   = {arXiv},
  eprint       = {2601.15621},
  bibsource    = {dblp computer science bibliography, https://dblp.org}
}

@article{DBLP:journals/corr/abs-2508-00828,
  author       = {Antoine Bigeard and
                  Langston Nashold and
                  Rayan Krishnan and
                  Shirley Wu},
  title        = {Finance Agent Benchmark: Benchmarking LLMs on Real-world Financial
                  Research Tasks},
  journal      = {CoRR},
  volume       = {abs/2508.00828},
  year         = {2025},
  url          = {https://doi.org/10.48550/arXiv.2508.00828},
  doi          = {10.48550/ARXIV.2508.00828},
  eprinttype   = {arXiv},
  eprint       = {2508.00828},
  bibsource    = {dblp computer science bibliography, https://dblp.org}
}

@article{phan2025lastexam,
      title = {A benchmark of expert-level academic questions to assess {AI} capabilities},
      author = {{Center for AI Safety} and {Scale AI} and {HLE Contributors Consortium}},
      journal = {Nature},
      volume = {649},
      pages = {1139--1146},
      year = {2026},
      doi = {10.1038/s41586-025-09962-4},
      eprint = {2501.14249},
      archivePrefix = {arXiv},
      primaryClass = {cs.LG},
      url = {https://arxiv.org/abs/2501.14249}
}

@inproceedings{DBLP:conf/iclr/QuDWCWY0W25,
  author       = {Changle Qu and
                  Sunhao Dai and
                  Xiaochi Wei and
                  Hengyi Cai and
                  Shuaiqiang Wang and
                  Dawei Yin and
                  Jun Xu and
                  Ji{-}Rong Wen},
  title        = {From Exploration to Mastery: Enabling LLMs to Master Tools via Self-Driven
                  Interactions},
  booktitle    = {The Thirteenth International Conference on Learning Representations,
                  {ICLR} 2025, Singapore, April 24-28, 2025},
  publisher    = {OpenReview.net},
  year         = {2025},
  url          = {https://openreview.net/forum?id=QKBu1BOAwd},
  bibsource    = {dblp computer science bibliography, https://dblp.org}
}

@inproceedings{fang2025play2prompt,
  title={Play2prompt: Zero-shot tool instruction optimization for llm agents via tool play},
  author={Fang, Wei and Zhang, Yang and Qian, Kaizhi and Glass, James and Zhu, Yada},
  booktitle={Findings of the Association for Computational Linguistics: ACL 2025},
  pages={26274--26290},
  year={2025}
}

@inproceedings{yuan2025easytool,
  title={Easytool: Enhancing llm-based agents with concise tool instruction},
  author={Yuan, Siyu and Song, Kaitao and Chen, Jiangjie and Tan, Xu and Shen, Yongliang and Ren, Kan and Li, Dongsheng and Yang, Deqing},
  booktitle={Proceedings of the 2025 Conference of the Nations of the Americas Chapter of the Association for Computational Linguistics: Human Language Technologies (Volume 1: Long Papers)},
  pages={951--972},
  year={2025}
}

@misc{trecmllm2025,
  author = {Kanoulas, Evangelos and Eustratiadis, Panagiotis and Sanderson, Mark and Callan, Jamie and Li, Yongkang and Qiao, Jingfen and Pal, Vaishali},
  title = {{TREC 2025 Million LLMs Track}},
  howpublished = {\url{https://trec-mllm.github.io/}},
  year = {2025},
}

@inproceedings{yue-etal-2025-masrouter,
    title = "{M}as{R}outer: Learning to Route {LLM}s for Multi-Agent Systems",
    author = "Yue, Yanwei  and
      Zhang, Guibin  and
      Liu, Boyang  and
      Wan, Guancheng  and
      Wang, Kun  and
      Cheng, Dawei  and
      Qi, Yiyan",
    editor = "Che, Wanxiang  and
      Nabende, Joyce  and
      Shutova, Ekaterina  and
      Pilehvar, Mohammad Taher",
    booktitle = "Proceedings of the 63rd Annual Meeting of the Association for Computational Linguistics (Volume 1: Long Papers)",
    month = jul,
    year = "2025",
    address = "Vienna, Austria",
    publisher = "Association for Computational Linguistics",
    url = "https://aclanthology.org/2025.acl-long.757/",
    doi = "10.18653/v1/2025.acl-long.757",
    pages = "15549--15572",
    ISBN = "979-8-89176-251-0"
}

@inproceedings{
zhang2025routerr,
title={Router-R1: Teaching {LLM}s Multi-Round Routing and Aggregation via Reinforcement Learning},
author={Haozhen Zhang and Tao Feng and Jiaxuan You},
booktitle={The Thirty-ninth Annual Conference on Neural Information Processing Systems},
year={2025},
url={https://openreview.net/forum?id=DWf4vroKWJ}
}

@article{DBLP:journals/corr/abs-2602-05366,
  author       = {Yichen Tang and
                  Weihang Su and
                  Yiqun Liu and
                  Qingyao Ai},
  title        = {Multi-Field Tool Retrieval},
  journal      = {CoRR},
  volume       = {abs/2602.05366},
  year         = {2026},
  url          = {https://doi.org/10.48550/arXiv.2602.05366},
  doi          = {10.48550/ARXIV.2602.05366},
  eprinttype   = {arXiv},
  eprint       = {2602.05366},
  bibsource    = {dblp computer science bibliography, https://dblp.org}
}

@inproceedings{craswell2021ms,
  title={Ms marco: Benchmarking ranking models in the large-data regime},
  author={Craswell, Nick and Mitra, Bhaskar and Yilmaz, Emine and Campos, Daniel and Lin, Jimmy},
  booktitle={Proceedings of the 44th international ACM SIGIR conference on research and development in information retrieval},
  pages={1566--1576},
  year={2021}
}

@article{craswell2025overview,
  title={Overview of the TREC 2022 deep learning track},
  author={Craswell, Nick and Mitra, Bhaskar and Yilmaz, Emine and Campos, Daniel and Lin, Jimmy and Voorhees, Ellen M and Soboroff, Ian},
  journal={arXiv preprint arXiv:2507.10865},
  year={2025}
}

@article{gu2024survey,
  title={A survey on llm-as-a-judge},
  author={Gu, Jiawei and Jiang, Xuhui and Shi, Zhichao and Tan, Hexiang and Zhai, Xuehao and Xu, Chengjin and Li, Wei and Shen, Yinghan and Ma, Shengjie and Liu, Honghao and others},
  journal={The Innovation},
  year={2024},
  publisher={Elsevier}
}

@inproceedings{wu2025natural,
  title={Natural Language Inference as a Judge: Detecting Factuality and Causality Issues in Language Model Self-Reasoning for Financial Analysis},
  author={Wu, Yilin and Yuan, Han and Zhang, Li and Ma, Zheng},
  booktitle={Proceedings of The 10th Workshop on Financial Technology and Natural Language Processing},
  pages={210--220},
  year={2025}
}

@article{li2025prorank,
  title={ProRank: Prompt Warmup via Reinforcement Learning for Small Language Models Reranking},
  author={Li, Xianming and Shakir, Aamir and Huang, Rui and Lipp, Julius and Li, Jing},
  journal={arXiv preprint arXiv:2506.03487},
  year={2025}
}

@article{qwen3embedding,
  title={Qwen3 Embedding: Advancing Text Embedding and Reranking Through Foundation Models},
  author={Zhang, Yanzhao and Li, Mingxin and Long, Dingkun and Zhang, Xin and Lin, Huan and Yang, Baosong and Xie, Pengjun and Yang, An and Liu, Dayiheng and Lin, Junyang and Huang, Fei and Zhou, Jingren},
  journal={arXiv preprint arXiv:2506.05176},
  year={2025}
}
\bibliographystyle{colm2026_conference}

\clearpage
\appendix
\section{More Details about \OURSBENCH{}}

\subsection{Schema Design}

A key challenge in constructing AgentBase lies in the heterogeneous representation of agents across platforms, where capability descriptions, usage instructions, and accessibility conditions are presented in inconsistent formats. To enable systematic indexing and fair comparison in agent search, we introduce a unified schema that standardizes agent information into four semantic groups:
(1) \textit{Agent metadata} provides stable identity and provenance signals, supporting deduplication and version tracking;
(2) \textit{Capability description} captures the functional semantics of agents through textual descriptions, category tags, and modality indicators;
(3) \textit{Usage guidance} characterizes how agents are invoked and interacted with in practice, including quick-start instructions and example interactions;
(4) \textit{Availability and constraints} record practical deployment conditions such as pricing, accessibility, base models, and update timestamps, reflecting real-world feasibility considerations.
This structured representation enables scalable agent collection, consistent retrieval over heterogeneous sources, and reproducible evaluation of agent search systems.
We show our schema design in \TAB~\ref{tab:schema} and one example in \TAB~\ref{tab:schema-example}.

\subsection{Example of Task Query and Task Description}

We show the example of single-and multi-agent task query, and task description in \TAB~\ref{tab:task-example}.

\subsection{Implementation Details of Benchmark Construction}

We use GPT-5.2 as the backbone for all generation steps with temperature $\tau = 1$. For task description generation, we begin with a candidate pool of 100 task queries, which are filtered down to 10 via rubric-based scoring across 5 criteria, each criterion associated with approximately 2 subtasks. For multi-agent task query we pick 2–4 subtasks (mean 2.91), sampled to be semantically related yet non-redundant. Concretely, given an anchor task query, we retrieve the top-$K$ most similar candidates and skip the highest-ranked $k_{\text{skip}}$ to avoid near-duplicate subtasks.

We retrieve a candidate set of top-$K$ agents using a hybrid scoring function
combining BM25~\citep{robertson2009probabilistic}, BGE~\citep{xiao2024cpack},
and ToolRet~\citep{shi2025retrieval} as lexical, semantic, and tool-aware
retrievers respectively:
\begin{align}
    s(a, \mathcal{T}_q) = \alpha\,s_{\text{lex}} + \beta\,s_{\text{sem}} + \gamma\,s_{\text{tool}}
\end{align}
where $\alpha + \beta + \gamma = 1$. Each component score is min-max normalised
before aggregation, and the top-$K$ agents are selected by the fused score.

\section{More Details about Experimental Setup}

\subsection{Completeness Computation}

For a given task $\mathcal{T}$ associated with $M \geq 1$ subtasks
$\{s_1, \dots, s_M\}$, each with ground-truth relevant set $\mathcal{R}_{s_m}$,
Completeness is defined as:
\begin{align}
    \text{COMP@}K(\mathcal{T}) = \mathbb{I}\!\left[
    \forall\, m \in \{1,\dots,M\} :
    \pi_K(\mathcal{T}) \cap \mathcal{R}_{s_m} \neq \emptyset
    \right],
\end{align}
where $\pi_K(\mathcal{T})$ denotes the top-$K$ retrieved tools.
A task is \textit{complete} at $K$ if the retrieved set contains at least
one relevant tool per subtask. For single-agent task query; when $M = 1$, Completeness reduces to
standard hit rate.

\begin{table}[t]
\centering
\small
\begin{tabular}{p{4cm} p{9cm}}
\toprule
\textbf{Schema Group} & \textbf{Representative Fields} \\
\midrule
Agent Metadata & Versioned agent ID, name, official URL, platform source \\
Capability Description & Functional description, category tags, supported modalities \\
Usage Guidance & Quick-start instructions, example input–output interactions \\
Availability \& Constraints & Pricing/accessibility, base model, update time, auxiliary metadata \\
\bottomrule
\end{tabular}
\caption{Unified schema for representing agents collected across platforms.}
\label{tab:schema}
\end{table}

\begin{table}[t]
\centering
\small
\renewcommand{\arraystretch}{1.3}
\begin{tabular}{ll}
\toprule
\textbf{Field} & \textbf{Value} \\
\midrule
ID          & \texttt{agt:openaiagents:2518c1@v1.1} \\
Source      & \texttt{openaiagents} \\
Name        & JAVA Code Guide \\
Description & A JAVA development assistant focusing on coding \\
            & standards and quality. \\
Tools       & \texttt{developer}, \texttt{browser}, \texttt{dalle}, \texttt{python} \\
Model       & GPT-5.2 \\
Input mode  & multi-modal \\
Quick start & \textit{Explain JAVA exception handling standards.} \\
            & \textit{How can I improve this MySQL query?} \\
            & \textit{Review my JAVA code snippet.} \\
            & \textit{What are the best practices for JAVA unit testing?} \\
Access      & free \\
URL         & https://chatgpt.com/g/g-EYiFThMtQ \\
Indexed     & 26 Jan 2026 \\
Misc & \begin{tabular}[t]{@{}p{0.62\linewidth}@{}}
         \texttt{gizmoId:}~\texttt{g-EYiFThMtQ};
         \texttt{gpt\_rating:}~4.0 \\[3pt]
         \small\textit{JAVA Code Guide is an AI-powered bot that leverages
         advanced GPT technology to assist JAVA developers in writing clean,
         efficient, and high-quality code. With its extensive knowledge of
         JAVA best practices~\dots}
       \end{tabular} \\
\bottomrule
\end{tabular}
\caption{An example agent from AgentBase.}
\label{tab:schema-example}
\end{table}

\begin{table}[t]
\centering
\small
\renewcommand{\arraystretch}{1.3}
\begin{tabular}{@{}lp{0.72\linewidth}@{}}
\toprule
\textbf{Type} & \textbf{Value} \\
\midrule
Task Description & Monitor, summarize, and compare news and web sources. \\[2pt]
Single-Agent Task Query & Create a mind map (hierarchical bullet nodes) summarizing
                   the Nature article at https://www.nature.com/articles/d41586-023-03988-2,
                   capturing its main sections, subsections, and key takeaways.
                   Assume I am a graduate student writing a quick briefing and
                   need a clear, structured outline. \\
Multi-Agent Task Query & I'm trying to reset my health and routine, so first ask me 5 questions about my schedule, priorities, and constraints, then build a 2-week productivity plan with a daily time-block schedule, a prioritized task list, and a simple tracker template, plus a 7-day Mediterranean meal plan at 1,800 kcal/day with recipes, step-by-step cooking tips, drink pairings, and a consolidated grocery list, and a 7-day English study plan with daily grammar drills, a vocabulary list, short speaking prompts, answer keys, and a simple progress tracker. \\
\bottomrule
\end{tabular}
\caption{An example Single-Agent Task Query, Multi-Agent Task Query, and a Task Description $T_d$ from AgentSearchBench.}
\label{tab:task-example}
\end{table}

\subsection{The Baselines}

We introduce more details of each type of model used in retrieval and reranking benchmark.

\paragraph{Retrieval} We evaluate four families of retrieval models:
(1)~\textit{Sparse}, based on term-weighting and learned sparse representations~\cite{robertson2009probabilistic, formal2021splade};
(2)~\textit{Dense}, using bi-encoder architectures trained for semantic similarity~\cite{santhanam2022colbertv2, izacard2021unsupervised, ni2022large, wang2021minilmv2, xiao2024cpack};
(3)~\textit{Tool-specific}, retrievers explicitly trained on tool corpora~\cite{qu2024towards, lu2025tools, zheng2403toolrerank, shi2025retrieval};
(4)~\textit{Decoder-only}, leveraging autoregressive LMs for retrieval~\cite{wang2024improving, DBLP:journals/corr/abs-2601-15621}.

\paragraph{Reranking} We evaluate four families of reranking models:
(1)~\textit{Cross-encoder}, scoring query--document pairs jointly~\cite{wang2021minilmv2, rerank2024mxbai, multi2024m3};
(2)~\textit{Tool-specific}, rerankers trained on tool corpora~\cite{zheng2403toolrerank};
(3)~\textit{Decoder-only}, reranking via autoregressive scoring~\cite{nogueira2020document, qwen3embedding};
(4)~\textit{LLM-based}, prompting large language models to produce relevance judgements~\cite{sun2023chatgpt}.

\subsection{The Prompts}

\begin{llmbox}{Prompt: Execution-Aware Probing Listwise Ranker}

\textbf{---SYSTEM---}\\[4pt]
You are an intelligent assistant that scores the quality of AI agent responses to a given probing query.\\[4pt]
I will provide you with \tvar{num} agent responses, each indicated by a number identifier \texttt{[i]}.
Score them based on their response quality to the provided probing query.\\[4pt]

\textbf{\#\# Scoring Guidelines (1--5)}
\begin{itemize}[noitemsep, topsep=2pt, leftmargin=*]
  \item \textbf{5 (Excellent):} Fully satisfies the user's request with correct, sufficient, and concise information.
  \item \textbf{4 (Good):} Mostly satisfies the request but has minor omissions or minor unnecessary detail.
  \item \textbf{3 (Fair):} Partially satisfies the request but is incomplete, indirect, or inefficient.
  \item \textbf{2 (Poor):} Minimally relevant or largely unhelpful.
  \item \textbf{1 (Very Poor):} Incorrect, irrelevant, or fails the task entirely.
\end{itemize}

\vspace{6pt}
\textbf{---USER---}\\[4pt]
\textbf{Probing Query:} \tvar{probe}\\[4pt]

\tvar{for i in range(num)}\\
\hspace*{1em}\texttt{[\tvar{i+1}]} \tvar{responses[i]}\\
\tvar{endfor}\\[4pt]

Score the \tvar{num} responses above based on their quality.\\[2pt]
The output format should be \texttt{[score[i], score[i+1], \dots]}, e.g., \texttt{[1, 5, 3, 5, \dots]} for each response. Only respond with the ordered scoring results.

\end{llmbox}

\section{More Results}

\subsection{Performance on Each Realistic Queries}

We show the results on the queries from two realistic benchmark in \TAB~\ref{tab:realistic_performance}, i.e., Humanity’s Last Exam (HLE) \cite{phan2025lastexam}, and Finance Agent Benchmark \cite{DBLP:journals/corr/abs-2508-00828}.

\subsection{More Results on Indexing}

\begin{table}[H]
\centering
\renewcommand{\arraystretch}{1.25}
\resizebox{\textwidth}{!}{%
\begin{tabular}{lccccccccccccc}
\toprule
\textbf{Model} & \textbf{Type} & \multicolumn{3}{c}{\textbf{NDCG}} & \multicolumn{3}{c}{\textbf{Precision}} & \multicolumn{3}{c}{\textbf{Recall}} & \multicolumn{3}{c}{\textbf{Completeness}}\\
\cline{3-14}
 & & \textbf{@10} & \textbf{@20} & \textbf{@50} & \textbf{@10} & \textbf{@20} & \textbf{@50} & \textbf{@10} & \textbf{@20} & \textbf{@50}  & \textbf{@10} & \textbf{@20} & \textbf{@50} \\
\midrule
\rowcolor{almond}\multicolumn{14}{c}{\textit{Task Description}} \\
\hline
BM25 & Sparse & 0.116 & 0.106 & 0.088 & 0.109 & 0.095 & 0.068 & 0.022 & 0.039 & 0.065 & 0.000 & 0.000 & 0.014 \\
SPLADE v2 & Sparse & 0.062 & 0.058 & 0.054 & 0.057 & 0.053 & 0.044 & 0.011 & 0.021 & 0.041 & 0.000 & 0.000 & 0.010 \\
\hline
ColBERT v2 & Dense & 0.143 & 0.139 & 0.136 & 0.136 & 0.131 & 0.112 & 0.028 & 0.053 & 0.115 & 0.010 & 0.019 & 0.058 \\
Contriever MS-Marco & Dense & 0.170 & 0.163 & 0.153 & 0.167 & 0.155 & 0.123 & 0.034 & 0.064 & 0.130 & 0.005 & 0.029 & 0.077 \\
GTR-T5 Base & Dense & 0.145 & 0.133 & 0.124 & 0.138 & 0.121 & 0.099 & 0.028 & 0.050 & 0.098 & 0.005 & 0.014 & 0.048 \\
MiniLM-L6 v2 & Dense & 0.152 & 0.151 & 0.139 & 0.155 & 0.148 & 0.114 & 0.035 & 0.067 & 0.119 & 0.010 & 0.029 & 0.053 \\
BGE-Large v1.5 & Dense & 0.186 & 0.171 & 0.160 & 0.182 & 0.160 & 0.129 & 0.039 & 0.066 & 0.134 & 0.000 & 0.024 & 0.062 \\
\hline
ToolRet & Tool & 0.139 & 0.134 & 0.124 & 0.142 & 0.131 & 0.103 & 0.029 & 0.054 & 0.103 & 0.000 & 0.019 & 0.043 \\
COLT ToolBench & Tool & 0.138 & 0.130 & 0.125 & 0.137 & 0.123 & 0.102 & 0.029 & 0.051 & 0.105 & 0.010 & 0.024 & 0.053 \\
COLT ToolLens & Tool & 0.088 & 0.082 & 0.077 & 0.088 & 0.077 & 0.063 & 0.018 & 0.031 & 0.062 & 0.000 & 0.010 & 0.024 \\
\hline
E5-Mistral 7B & Decoder-only & 0.025 & 0.025 & 0.026 & 0.027 & 0.026 & 0.023 & 0.005 & 0.009 & 0.023 & 0.000 & 0.000 & 0.005 \\
Qwen-Embedding 8B & Decoder-only & 0.174 & 0.164 & 0.153 & 0.164 & 0.152 & 0.124 & 0.032 & 0.063 & 0.127 & 0.005 & 0.019 & 0.062 \\
\bottomrule
\end{tabular}%
}
\caption{Retrieval results on Task Description $T_d$ with \textbf{full} indexing.}
\label{tab:retrieval_app_desc}
\end{table}

\begin{table}[H]
\centering
\renewcommand{\arraystretch}{1.25}
\resizebox{\textwidth}{!}{%
\begin{tabular}{lccccccccccccc}
\toprule
\textbf{Model} & \textbf{Type} & \multicolumn{3}{c}{\textbf{NDCG}} & \multicolumn{3}{c}{\textbf{Precision}} & \multicolumn{3}{c}{\textbf{Recall}} & \multicolumn{3}{c}{\textbf{Hit}}\\
\cline{3-14}
 & & \textbf{@1} & \textbf{@5} & \textbf{@20} & \textbf{@1} & \textbf{@5} & \textbf{@20} & \textbf{@1} & \textbf{@5} & \textbf{@20} & \textbf{@1} & \textbf{@5} & \textbf{@20} \\
\midrule
\rowcolor{lavenderblush}\multicolumn{14}{c}{\textit{Single-Agent Task Query}} \\
\hline
BM25 & Sparse & 0.332 & 0.297 & 0.309 & 0.332 & 0.269 & 0.158 & 0.038 & 0.157 & 0.366 & 0.332 & 0.669 & 0.853 \\
SPLADE v2 & Sparse & 0.096 & 0.081 & 0.091 & 0.096 & 0.072 & 0.048 & 0.011 & 0.044 & 0.114 & 0.096 & 0.256 & 0.484 \\
\hline
ColBERT v2 & Dense & 0.317 & 0.275 & 0.295 & 0.317 & 0.248 & 0.154 & 0.036 & 0.141 & 0.356 & 0.317 & 0.629 & 0.830 \\
Contriever MS-Marco & Dense & 0.323 & 0.273 & 0.293 & 0.323 & 0.245 & 0.154 & 0.037 & 0.140 & 0.351 & 0.323 & 0.633 & 0.835 \\
GTR-T5 Base & Dense & 0.296 & 0.243 & 0.253 & 0.296 & 0.216 & 0.129 & 0.034 & 0.124 & 0.296 & 0.296 & 0.588 & 0.792 \\
MiniLM-L6 v2 & Dense & 0.327 & 0.272 & 0.284 & 0.327 & 0.244 & 0.147 & 0.035 & 0.136 & 0.335 & 0.327 & 0.638 & 0.820 \\
BGE-Large v1.5 & Dense & 0.371 & 0.330 & 0.362 & 0.371 & 0.299 & 0.192 & 0.042 & 0.173 & 0.438 & 0.371 & 0.692 & 0.878 \\
\hline
ToolRet & Tool & 0.322 & 0.283 & 0.312 & 0.322 & 0.259 & 0.167 & 0.034 & 0.144 & 0.382 & 0.322 & 0.656 & 0.863 \\
COLT ToolBench & Tool & 0.304 & 0.241 & 0.251 & 0.304 & 0.211 & 0.128 & 0.036 & 0.121 & 0.293 & 0.304 & 0.581 & 0.788 \\
COLT ToolLens & Tool & 0.234 & 0.191 & 0.200 & 0.234 & 0.170 & 0.105 & 0.026 & 0.096 & 0.233 & 0.234 & 0.506 & 0.712 \\
\hline
E5-Mistral 7B & Decoder-only & 0.079 & 0.063 & 0.068 & 0.079 & 0.057 & 0.036 & 0.008 & 0.031 & 0.082 & 0.079 & 0.215 & 0.393 \\
Qwen-Embedding 8B & Decoder-only & 0.297 & 0.238 & 0.259 & 0.297 & 0.212 & 0.136 & 0.034 & 0.118 & 0.312 & 0.297 & 0.571 & 0.788 \\
\bottomrule
\end{tabular}%
}
\caption{Retrieval results on Single-Agent Task Query $T_q$ with \textbf{full} indexing.}
\label{tab:retrieval_app_single_agent}
\end{table}

\begin{table}[H]
\centering
\renewcommand{\arraystretch}{1.25}
\resizebox{\textwidth}{!}{%
\begin{tabular}{lcccccccccccccccc}
\toprule
\textbf{Model} & \textbf{Type} & \multicolumn{3}{c}{\textbf{NDCG}} & \multicolumn{3}{c}{\textbf{Precision}} & \multicolumn{3}{c}{\textbf{Recall}} & \multicolumn{3}{c}{\textbf{Hit}} & \multicolumn{3}{c}{\textbf{Completeness}}\\
\cline{3-17}
 & & \textbf{@1} & \textbf{@5} & \textbf{@20} & \textbf{@1} & \textbf{@5} & \textbf{@20} & \textbf{@1} & \textbf{@5} & \textbf{@20} & \textbf{@1} & \textbf{@5} & \textbf{@20} & \textbf{@1} & \textbf{@5} & \textbf{@20} \\
\midrule
\rowcolor{lemonchiffon}\multicolumn{17}{c}{\textit{Multi-Agent Task Query}} \\
\hline
BM25 & Sparse & 0.318 & 0.257 & 0.189 & 0.318 & 0.238 & 0.146 & 0.015 & 0.054 & 0.129 & 0.318 & 0.654 & 0.886 & 0.012 & 0.074 & 0.244 \\
SPLADE v2 & Sparse & 0.102 & 0.080 & 0.066 & 0.102 & 0.074 & 0.053 & 0.005 & 0.017 & 0.048 & 0.102 & 0.298 & 0.594 & 0.002 & 0.012 & 0.050 \\
\hline
ColBERT v2 & Dense & 0.326 & 0.251 & 0.187 & 0.326 & 0.232 & 0.147 & 0.015 & 0.050 & 0.129 & 0.326 & 0.618 & 0.822 & 0.004 & 0.034 & 0.078 \\
Contriever MS-Marco & Dense & 0.254 & 0.225 & 0.185 & 0.254 & 0.217 & 0.156 & 0.012 & 0.047 & 0.137 & 0.254 & 0.610 & 0.860 & 0.014 & 0.070 & 0.194 \\
GTR-T5 Base & Dense & 0.282 & 0.225 & 0.182 & 0.282 & 0.211 & 0.152 & 0.012 & 0.045 & 0.128 & 0.282 & 0.570 & 0.836 & 0.010 & 0.048 & 0.168 \\
MiniLM-L6 v2 & Dense & 0.304 & 0.256 & 0.205 & 0.304 & 0.241 & 0.169 & 0.014 & 0.052 & 0.147 & 0.304 & 0.652 & 0.882 & 0.008 & 0.060 & 0.214 \\
BGE-Large v1.5 & Dense & 0.378 & 0.310 & 0.250 & 0.378 & 0.292 & 0.207 & 0.016 & 0.063 & 0.180 & 0.378 & 0.714 & 0.926 & 0.010 & 0.076 & 0.232 \\
\hline
ToolRet & Tool & 0.300 & 0.269 & 0.224 & 0.300 & 0.258 & 0.190 & 0.013 & 0.057 & 0.164 & 0.300 & 0.682 & 0.914 & 0.006 & 0.094 & 0.270 \\
COLT ToolBench & Tool & 0.230 & 0.191 & 0.157 & 0.230 & 0.180 & 0.130 & 0.010 & 0.039 & 0.116 & 0.230 & 0.536 & 0.818 & 0.010 & 0.054 & 0.154 \\
COLT ToolLens & Tool & 0.194 & 0.166 & 0.140 & 0.194 & 0.157 & 0.118 & 0.008 & 0.035 & 0.102 & 0.194 & 0.502 & 0.772 & 0.006 & 0.040 & 0.144 \\
\hline
E5-Mistral 7B & Decoder-only & 0.012 & 0.018 & 0.018 & 0.012 & 0.020 & 0.017 & 0.000 & 0.004 & 0.015 & 0.012 & 0.092 & 0.262 & 0.000 & 0.000 & 0.010 \\
Qwen-Embedding 8B & Decoder-only & 0.228 & 0.201 & 0.175 & 0.228 & 0.193 & 0.149 & 0.011 & 0.042 & 0.133 & 0.228 & 0.566 & 0.842 & 0.010 & 0.046 & 0.178 \\
\bottomrule
\end{tabular}%
}
\caption{Retrieval results on Multi-Agent Task Query $T_q$ with \textbf{full} indexing.}
\label{tab:retrieval_app_multi_agent}
\end{table}

\begin{table}[H]
\centering
\renewcommand{\arraystretch}{1.25}
\resizebox{\textwidth}{!}{%
\begin{tabular}{lcccccccccc}
\toprule
\textbf{Model} & \textbf{Type} & \multicolumn{3}{c}{\textbf{NDCG}} & \multicolumn{3}{c}{\textbf{Precision}} & \multicolumn{3}{c}{\textbf{Recall}} \\
\cline{3-11}
 & & \textbf{@1} & \textbf{@5} & \textbf{@20} & \textbf{@1} & \textbf{@5} & \textbf{@20} & \textbf{@1} & \textbf{@5} & \textbf{@20} \\
\midrule
\rowcolor{mintcream}\multicolumn{11}{c}{\textit{HLE: Humanity's Last Exam}} \\
\hline
\textbf{BM25} & Sparse & 0.042 & 0.022 & 0.028 & 0.042 & 0.017 & 0.011 & 0.006 & 0.013 & 0.028 \\
\textbf{SPLADE v2} & Sparse & 0.021 & 0.013 & 0.025 & 0.021 & 0.008 & 0.012 & 0.003 & 0.014 & 0.036 \\
\textbf{ColBERT v2} & Dense & 0.104 & 0.066 & 0.071 & 0.104 & 0.050 & 0.026 & 0.022 & 0.039 & 0.080 \\
\textbf{BGE-large v1.5} & Dense & 0.208 & 0.191 & 0.194 & 0.208 & 0.150 & 0.076 & 0.020 & 0.163 & 0.258 \\
\textbf{E5 Mistral 7B} & Decoder-only & 0.042 & 0.057 & 0.070 & 0.042 & 0.058 & 0.032 & 0.007 & 0.040 & 0.098 \\
\textbf{ToolRet} & Tool & 0.250 & 0.194 & 0.225 & 0.250 & 0.150 & 0.089 & 0.033 & 0.137 & 0.301 \\
\textbf{Tool-Embed} & Tool & 0.271 & 0.181 & 0.174 & 0.271 & 0.142 & 0.060 & 0.043 & 0.116 & 0.181 \\
\midrule
\rowcolor{seashell}\multicolumn{11}{c}{\textit{Finance Agent Benchmark}} \\
\hline
\textbf{BM25} & Sparse & 0.061 & 0.047 & 0.049 & 0.061 & 0.042 & 0.020 & 0.009 & 0.024 & 0.053 \\
\textbf{SPLADE v2} & Sparse & 0.000 & 0.050 & 0.067 & 0.000 & 0.048 & 0.024 & 0.000 & 0.046 & 0.120 \\
\textbf{ColBERT v2} & Dense & 0.030 & 0.043 & 0.044 & 0.030 & 0.036 & 0.017 & 0.003 & 0.033 & 0.048 \\
\textbf{BGE-large v1.5} & Dense & 0.182 & 0.166 & 0.233 & 0.182 & 0.115 & 0.065 & 0.046 & 0.157 & 0.365 \\
\textbf{E5 Mistral 7B} & Decoder-only & 0.121 & 0.125 & 0.138 & 0.121 & 0.085 & 0.039 & 0.035 & 0.106 & 0.156 \\
\textbf{ToolRet} & Tool & 0.242 & 0.147 & 0.198 & 0.242 & 0.103 & 0.068 & 0.058 & 0.087 & 0.266 \\
\textbf{Tool-Embed} & Tool & 0.212 & 0.116 & 0.117 & 0.212 & 0.079 & 0.030 & 0.048 & 0.078 & 0.111 \\
\bottomrule
\end{tabular}%
}
\caption{Retrieval results on 200 Real Task Query $T_q$.}
\label{tab:realistic_performance}
\end{table}

\begin{table}[H]
\centering
\renewcommand{\arraystretch}{1.25}
\resizebox{\textwidth}{!}{%
\begin{tabular}{lccccccccccc}
\toprule
\textbf{Model} & \textbf{Type} & \multicolumn{3}{c}{\textbf{NDCG}} & \multicolumn{3}{c}{\textbf{Completeness}} & \textbf{MRR} & \textbf{$\rho$} &
\textbf{$\tau$} & \textbf{RBO@0.9}\\
\cline{3-12}
 & & \textbf{@1} & \textbf{@5} & \textbf{@20} & \textbf{@1} & \textbf{@5} & \textbf{@20} & & & & \\
\midrule
\rowcolor{aliceblue}\multicolumn{12}{c}{\textit{Task Description}} \\
\hline
MiniLM-L12 v2 & Cross-Encoder & 0.580 & 0.686 & 0.842 & 0.000 & 0.000 & 0.200 & 0.218 & 0.280 & 0.229 & 0.397 \\
BGE Reranker v2 & Cross-Encoder & 0.530 & 0.659 & 0.831 & 0.000 & 0.000 & 0.200 & 0.185 & 0.283 & 0.223 & 0.400 \\
MXBAI Reranker Large & Cross-Encoder & 0.480 & 0.555 & 0.796 & 0.000 & 0.000 & 0.200 & 0.215 & 0.185 & 0.149 & 0.415 \\
\hline
Tool-Rank 4B & Tool & 0.580 & 0.652 & 0.840 & 0.000 & 0.000 & 0.200 & 0.108 & 0.268 & 0.209 & 0.417 \\
Tool-Rank 8B & Tool & 0.440 & 0.642 & 0.810 & 0.000 & 0.000 & 0.200 & 0.134 & 0.184 & 0.138 & 0.372 \\
\hline
MonoT5 Base MS-Marco & Decoder-only & 0.480 & 0.639 & 0.806 & 0.000 & 0.000 & 0.200 & 0.164 & 0.115 & 0.083 & 0.380 \\
Qwen Reranker 0.6B & Decoder-only & 0.580 & 0.642 & 0.828 & 0.000 & 0.000 & 0.200 & 0.157 & 0.205 & 0.158 & 0.409 \\
\textit{Qwen Reranker 4B} (payload trunc.) & Decoder-only & 0.580 & 0.669 & 0.835 & 0.000 & 0.000 & 0.200 & 0.126 & 0.279 & 0.216 & 0.411 \\
\hline
RankGPT GPT-5.2 & LLM & 0.480 & 0.641 & 0.833 & 0.000 & 0.000 & 0.200 & 0.133 & 0.403 & 0.322 & 0.424 \\
RankGPT LLaMA-3.3 70B & LLM & 0.380 & 0.604 & 0.813 & 0.000 & 0.000 & 0.200 & 0.125 & 0.264 & 0.210 & 0.421 \\
RankGPT Qwen-3 32B & LLM & 0.580 & 0.608 & 0.820 & 0.000 & 0.000 & 0.200 & 0.142 & 0.260 & 0.202 & 0.393 \\
\hline
Random Shuffle & Proxy & 0.410 & 0.497 & 0.771 & 0.000 & 0.000 & 0.200 & 0.105 & 0.095 & 0.073 & 0.350 \\
\bottomrule
\end{tabular}%
}
\caption{Reranking results on Task Description $T_d$ with \textbf{golden labels} and \textbf{full} indexing.}
\label{tab:reranking_app_desc}
\end{table}

\begin{table}[H]
\centering
\renewcommand{\arraystretch}{1.25}
\resizebox{\textwidth}{!}{%
\begin{tabular}{lccccccccccc}
\toprule
\textbf{Model} & \textbf{Type} & \multicolumn{3}{c}{\textbf{NDCG}} & \multicolumn{3}{c}{\textbf{Completeness}} & \textbf{MRR} & \textbf{$\rho$} &
\textbf{$\tau$} & \textbf{RBO@0.9}\\
\cline{3-12}
 & & \textbf{@1} & \textbf{@5} & \textbf{@20} & \textbf{@1} & \textbf{@5} & \textbf{@20} & & & & \\
\midrule
\rowcolor{aliceblue}\multicolumn{12}{c}{\textit{Task Query}} \\
MiniLM-L12 v2 & Cross-Encoder & 0.535 & 0.530 & 0.765 & 0.300 & 0.545 & 0.850 & 0.360 & 0.040 & 0.030 & 0.450 \\
BGE Reranker v2 & Cross-Encoder & 0.650 & 0.630 & 0.810 & 0.305 & 0.600 & 0.850 & 0.340 & 0.115 & 0.110 & 0.470 \\
MXBAI Reranker Large & Cross-Encoder & 0.480 & 0.515 & 0.730 & 0.290 & 0.585 & 0.850 & 0.300 & 0.110 & 0.090 & 0.440 \\
\hline
Tool-Rank 4B & Tool & 0.655 & 0.665 & 0.830 & 0.305 & 0.620 & 0.850 & 0.380 & 0.260 & 0.215 & 0.490 \\
Tool-Rank 8B & Tool & 0.700 & 0.655 & 0.825 & 0.325 & 0.605 & 0.850 & 0.330 & 0.250 & 0.210 & 0.480 \\
\hline
MonoT5 Base MS-Marco & Decoder-only & 0.550 & 0.575 & 0.785 & 0.280 & 0.610 & 0.850 & 0.340 & 0.120 & 0.100 & 0.450 \\
Qwen Reranker 0.6B & Decoder-only & 0.650 & 0.645 & 0.820 & 0.310 & 0.595 & 0.850 & 0.360 & 0.220 & 0.180 & 0.475 \\
Qwen Reranker 4B & Decoder-only & 0.660 & 0.655 & 0.825 & 0.305 & 0.610 & 0.850 & 0.360 & 0.250 & 0.210 & 0.485 \\
\hline
RankGPT GPT-5.2 & LLM & 0.650 & 0.670 & 0.825 & 0.300 & 0.610 & 0.850 & 0.365 & 0.245 & 0.205 & 0.485 \\
RankGPT LLaMA-3.3 70B & LLM & 0.580 & 0.610 & 0.795 & 0.260 & 0.515 & 0.850 & 0.355 & 0.165 & 0.135 & 0.470 \\
RankGPT Qwen-3 32B & LLM & 0.605 & 0.620 & 0.790 & 0.295 & 0.550 & 0.850 & 0.360 & 0.165 & 0.140 & 0.460 \\
\hline
Random Shuffle & Proxy & 0.460 & 0.500 & 0.740 & 0.280 & 0.665 & 0.850 & 0.205 & 0.030 & 0.025 & 0.370 \\
\bottomrule
\end{tabular}%
}
\caption{Reranking results on Task Query $T_q$ (Single-Agent and Multi-Agent Task Queries) with \textbf{golden labels} and \textbf{full} indexing.}
\label{tab:reranking_app_query}
\end{table}

\end{document}